\documentclass[final,5p,times,twocolumn,authoryear]{elsarticle}
\pdfoutput=1

\usepackage{times}
\usepackage[hyphens]{url}
\usepackage{breakcites} %
\usepackage{color}
\usepackage{float}
\usepackage{amssymb}
\usepackage{amsmath}
\usepackage{graphicx}
\usepackage{tikz}
\usepackage{scalefnt}
\usepackage[mode=buildnew]{standalone}
\usepackage{bm}
\usepackage{enumitem}
\usepackage{booktabs}
\usepackage{multirow}
\usepackage{adjustbox}
\usepackage{subcaption}

\def\eqref#1{equation~\ref{#1}}

\def\1{\bm{1}}

\def\va{{\bm{a}}}
\def\vb{{\bm{b}}}
\def\vc{{\bm{c}}}
\def\vd{{\bm{d}}}

\def\vh{{\bm{h}}}
\def\vi{{\bm{i}}}

\def\vk{{\bm{k}}}
\def\vl{{\bm{l}}}

\def\vo{{\bm{o}}}

\def\vr{{\bm{r}}}
\def\vs{{\bm{s}}}

\def\vv{{\bm{v}}}

\def\vx{{\bm{x}}}

\def\vz{{\bm{z}}}
\def\v#1{{\bm{#1}}}

\def\mA{{\bm{A}}}

\def\mM{{\bm{M}}}

\def\mW{{\bm{W}}}

\def\m#1{{\bm{#1}}}

\DeclareMathAlphabet{\mathsfit}{\encodingdefault}{\sfdefault}{m}{sl}
\SetMathAlphabet{\mathsfit}{bold}{\encodingdefault}{\sfdefault}{bx}{n}

\def\sR{{\mathbb{R}}}

\newcommand{\Ls}{\mathcal{L}}

\newcommand{\softmax}{\mathrm{softmax}}
\newcommand{\sigmoid}{\sigma}

\DeclareMathOperator*{\argmax}{arg\,max}

\usepackage{soul}

\newcommand{\specialcell}[2][l]{\begin{tabular}[#1]{@{}l@{}}#2\end{tabular}}

\definecolor{darkgreen}{RGB}{0,128,0}

\usepackage[symbols,nogroupskip,sort=none,acronym]{glossaries-extra}
\glsdisablehyper
\newacronym{vg}{VG}{visual grounding}
\newacronym{rpn}{RPN}{Region Proposal Network}
\newacronym{pipeline}{URS}{Uncertainty Resolving System}
\newacronym{reg_model}{A-REG}{Attribute-Referring Expression Generator}
\newacronym{mmi-mm}{MMI-MM}{Max-Margin Maximum Mutual Information}
\glsxtrnewsymbol[description={The command}]{m:com}{\ensuremath{c}}
\glsxtrnewsymbol[description={The Image}]{m:img}{\ensuremath{I}}
\glsxtrnewsymbol[description={Set of objects in the image \gls{m:img}}]{m:objs}{\ensuremath{O_{\gls{m:img}}}}
\glsxtrnewsymbol[description={The number of objects in \gls{m:objs}}]{m:num_obj}{\ensuremath{k}}
\glsxtrnewsymbol[description={One object $\in \gls{m:objs}$}]{m:obj}{\ensuremath{o}}
\glsxtrnewsymbol[description={The target object}]{m:t-obj}{\ensuremath{o^*}}
\glsxtrnewsymbol[description={Temperature Scaling hyper-parameter}]{m:tau}{\ensuremath{\tau}}
\glsxtrnewsymbol[description={The input variables of the model}]{m:vars}{\ensuremath{\Phi}}
\glsxtrnewsymbol[description={Probability distribution over all objects in image \gls{m:img}}]{m:obj-dist}{\ensuremath{p(\gls{m:objs} \vert \gls{m:vars}, \m{\theta})}}
\glsxtrnewsymbol[description={The parameters for a model}]{m:par}{\ensuremath{\theta}}
\glsxtrnewsymbol[description={Current hidden state of first LSTM layer}]{m:ht1}{\ensuremath{\vh_1^{(t)}}}
\glsxtrnewsymbol[description={Current hidden state of second LSTM layer}]{m:ht2}{\ensuremath{\vh_2^{(t)}}}
\glsxtrnewsymbol[description={Previous hidden state of first LSTM layer}]{m:ht1-1}{\ensuremath{\vh_1^{(t-1)}}}
\glsxtrnewsymbol[description={Previous hidden state of second LSTM layer}]{m:ht2-1}{\ensuremath{\vh_2^{(t-1)}}}
\glsxtrnewsymbol[description={Input vector of first LSTM layer}]{m:i1}{\ensuremath{\vi_1}}
\glsxtrnewsymbol[description={Input vector of second LSTM layer}]{m:i2}{\ensuremath{\vi_2}}
\glsxtrnewsymbol[description={The feature vector of the object \gls{m:obj}}]{m:f-obj}{\ensuremath{\vo}}
\glsxtrnewsymbol[description={The vector size for \gls{m:f-obj}}]{m:f-obj-dim}{\ensuremath{\gls{m:f-obj}\in\sR^{2048}}}
\glsxtrnewsymbol[description={The image feature vector}]{m:f-img}{\ensuremath{\vv_{\gls{m:img}}}}
\glsxtrnewsymbol[description={The vector size for \gls{m:f-img}}]{m:f-img-dim}{\ensuremath{\gls{m:f-img}\in\sR^{2048}}}
\glsxtrnewsymbol[description={The action attribute feature vector}]{m:f-act}{\ensuremath{\va_{\gls{m:img}}}}
\glsxtrnewsymbol[description={The color attribute feature vector}]{m:f-col}{\ensuremath{\vc_{\gls{m:img}}}}
\glsxtrnewsymbol[description={The location attribute feature vector}]{m:f-loc}{\ensuremath{\vl_{\gls{m:img}}}}
\glsxtrnewsymbol[description={The dimension of \gls{m:f-act}}]{m:f-act-dim}{\ensuremath{\gls{m:f-act}\in\sR^{11}}}
\glsxtrnewsymbol[description={The dimension of \gls{m:f-col}}]{m:f-col-dim}{\ensuremath{\gls{m:f-col}\in\sR^{12}}}
\glsxtrnewsymbol[description={The dimension of \gls{m:f-loc}}]{m:f-loc-dim}{\ensuremath{\gls{m:f-loc}\in\sR^{3}}}
\glsxtrnewsymbol[description={The bounding box feature vector}]{m:f-box}{\ensuremath{\vb_{\gls{m:img}}}}
\glsxtrnewsymbol[description={The class label feature vector}]{m:f-cls}{\ensuremath{\vk_{\gls{m:img}}}}
\glsxtrnewsymbol[description={The dimension of \gls{m:f-cls}}]{m:f-cls-dim}{\ensuremath{\vk_{\gls{m:img}}\in\sR^{23}}}
\glsxtrnewsymbol[description={The distance count feature vector}]{m:f-cnt}{\ensuremath{\vd_{\gls{m:img}}}}
\glsxtrnewsymbol[description={The dimension of \gls{m:f-cnt}}]{m:f-cnt-dim}{\ensuremath{\vd_{\gls{m:img}}\in\sR^{6}}}
\glsxtrnewsymbol[description={The attributes matrix}]{m:m-attr}{\ensuremath{\mA_{\gls{m:img}}}}
\glsxtrnewsymbol[description={The weighted combined attributes feature vector}]{m:f-attr}{\ensuremath{\va_{\gls{m:img}}}}
\glsxtrnewsymbol[description={The switch deciding which predictor to use}]{m:target_switch}{\ensuremath{\gamma_{\gls{m:img}}^{(t)}}}
\glsxtrnewsymbol[description={The predicted switch value}]{m:switch}{\ensuremath{\hat{\gamma}_{\gls{m:img}}^{(t)}}}
\glsxtrnewsymbol[description={Weight for the \acrshort{mmi-mm} loss}]{m:weight_mmi}{\ensuremath{\lambda_{MMI}}}
\glsxtrnewsymbol[description={Weight for the switch loss}]{m:weight_switch}{\ensuremath{\lambda_{switch}}}
\glsxtrnewsymbol[description={The generated expression}]{m:expr}{\ensuremath{\hat{X}}}
\glsxtrnewsymbol[description={The target expression}]{m:target_expr}{\ensuremath{X}}
\glsxtrnewsymbol[description={One word at timestep $t$ of generated expression}]{m:expr_t}{\ensuremath{x^{(t)}}}
\glsxtrnewsymbol[description={One word at timestep $t$ of generated expression}]{m:expr_t-1}{\ensuremath{x^{(t-1)}}}
\glsxtrnewsymbol[description={The target expression}]{m:target_expr_t}{\ensuremath{\hat{x}^{(t)}}}
\glsxtrnewsymbol[description={relative increase meteor}]{m:meteor-relative}{6\%}
\glsxtrnewsymbol[description={relative increase rouge}]{m:rouge-relative}{8\%}
\glsxtrnewsymbol[description={absolute increase rouge}]{m:ambiguous-absolute-increase}{12.6\%}
\glsxtrnewsymbol[description={absolute increase rouge}]{m:overall-increase-VG}{9\%}
\graphicspath{{images/}}
\journal{Engineering Applications of Artificial Intelligence}
\begin{document}
\begin{frontmatter}
\title{
Giving Commands to a Self-Driving Car: How to Deal with Uncertain Situations?}
\author[kul]{Thierry Deruyttere\corref{t2}}
\author[kul]{Victor Milewski\corref{t1}}
\author[kul]{Marie-Francine Moens}
\cortext[t1]{Equal contribution}
\cortext[t2]{Equal contribution and corresponding author.}
\address[kul]{KU Leuven, 200A Celestijnenlaan, Leuven, 3001 Belgium\\\{thierry.deruyterre ; victor.milewski ; sien.moens\} @kuleuven.be}
\begin{abstract}
Current technology for autonomous cars primarily focuses on getting the passenger from point A to B. Nevertheless, it has been shown that passengers are afraid of taking a ride in self-driving cars.
One way to alleviate this problem is by allowing the passenger to give natural language commands to the car.
However, the car can misunderstand the issued command or the visual surroundings which could lead to uncertain situations.
It is desirable that the self-driving car detects these situations and interacts with the passenger to solve them.
This paper proposes a model that detects the uncertain situations when a command is given and finds the visual objects causing it.
Optionally, a question generated by the system describing the uncertain objects is included.
We argue that if the car could explain the objects in a human-like way, passengers could gain more confidence in the car's abilities.
Thus, we investigate how to (1) detect uncertain situations and their underlying causes, and (2) how to generate clarifying questions for the passenger.
When evaluating on the Talk2Car dataset, we show that the proposed model, \acrfull{pipeline}, improves \gls{m:ambiguous-absolute-increase} in terms of $IoU_{.5}$ compared to not using \gls{pipeline}.
Furthermore, we designed a referring expression generator (REG) \acrfull{reg_model} tailored to a self-driving car setting which yields a relative improvement of \gls{m:meteor-relative} METEOR and \gls{m:rouge-relative} ROUGE-l compared with state-of-the-art REG models, and is three times faster.

\end{abstract}
\begin{keyword}
Self-Driving Cars \sep Question Generation \sep Uncertainty Detection
\end{keyword}
\end{frontmatter}
\section{Introduction}
\label{sect:introduction}
\glsresetall
\begin{figure}[htp]
  \centering
    \includegraphics[width=.95\linewidth]{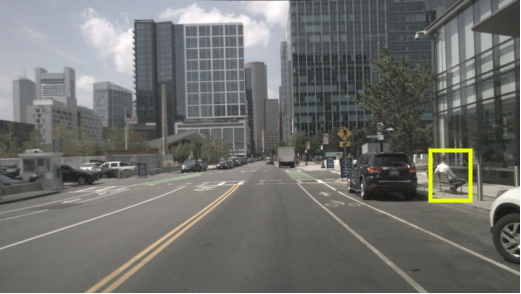}
\caption{Example from the Talk2Car dataset.}{ \textit{Command}: ``There is \textbf{mark on the bench}! Pull over here and park so I can go grab lunch with him''. The referred object is indicated with the yellow bounding box in the image and in bold font in the text. Best viewed in color.}
\label{fig:t2c-example}
\end{figure}
\begin{figure*}[th]
  \centering
    \includegraphics[width=\textwidth]{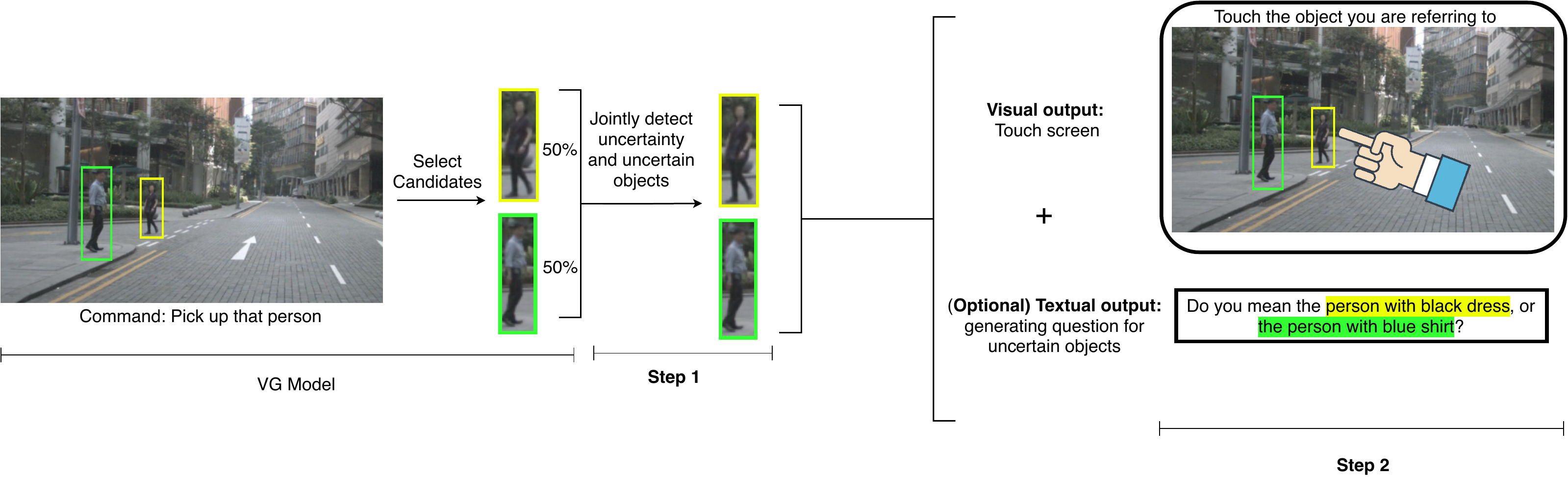}
    \caption{The full \acrfull{pipeline} on top of a \acrfull{vg} model.}{The \acrfull{pipeline} works by receiving the output of any \acrfull{vg} model that outputs a softmax distribution.
    The \gls{vg} model is given an image \gls{m:img}, a command \gls{m:com} (for instance: ``Pick up that person''), and a set of objects \gls{m:objs} (in this case, they are limited to the green and yellow bounding box) as inputs.
    Based on these outputs, it makes a prediction to which object is being referred.
    In \textbf{Step 1} of \gls{pipeline}, we detect, based on the output of the \gls{vg} model, if there is an ambiguous situation and also which objects are causing this uncertainty.
    For \textbf{Step 2} of \gls{pipeline}, we have a \textbf{visual} output and optionally also a \textbf{textual} output.
    In the \textbf{visual} output, we display the uncertain objects as an overlay on the image, for instance, on a touch screen.
    The user can then indicate which object they are referring to.
    We can also have an additional \textbf{textual} output that shows a question, explaining the uncertain objects to the passenger. Best viewed in color.
    }
\label{fig:proposed_system}
\end{figure*}
In the last few years, we have witnessed a fast-growing interest in self-driving cars. Many researchers are involved in creating the first completely autonomous vehicle that drives in open realistic environments where a human driver becomes obsolete.
However, surveys~\citep{howard2014public,schoettle2014survey,PublicPerceptionSDC,AAAFearOfSDC}
show that people might feel uneasy or even scared to sit in a self-driving car where they no longer have control over the car's actions.
For this study, we also held a survey on several social media platforms asking how much people trust self-driving cars. Out of the 254 responses, 36.8\% indicated not to trust driving around in a self-driving car. As reported in previous surveys, the majority of these people explained to be afraid of not being in control anymore or to have safety concerns about autonomous cars.
Furthermore, we noted that with the ability to give commands to self-driving cars, the number decreased to 30.4\%.

This is in line with recent studies, in which datasets are proposed to give such commands to self-driving cars.
Examples of such datasets are Talk2Car~\citep{deruyttere2019talk2car}, Talk2Nav~\citep{vasudevan2019talk2nav}, and Touchdown~\citep{chen2018touchdown}, to name a few.
The former investigates commands that should be executed in the car's visual field.
An example of such a command is visible in Figure~\ref{fig:t2c-example}).
The latter two investigate natural language navigation commands that contain directional guidance (i.e., ``cross the intersection and then turn left, when you see a red building turn right'') and require the car to drive through the city to achieve its end goal.

To the best of our knowledge, all studies ignore that the issued command and the visual scene might result in an uncertain or ambiguous interpretation by the system.
We believe that it is essential for a system to recognize and resolve these ambiguous situations.
An example of this is given in the Talk2Car setting.
This dataset includes commands such as ``Stop next to this guy with his blue shirt. I need to pick him up. He is my friend''. Nevertheless, suppose multiple persons are wearing a blue shirt. In that case, it might not be apparent to the car which person is being referred to.
We hypothesize that resolving these ambiguous situations can be achieved in two ways.
Either the car displays the source of ambiguity on a screen or, in addition to the former, the car can ask confirming questions to the passenger in either textual form or by speech.
In the example, the car could generate a clarifying question such as ``Do you mean the man on the left or the man on the right?'' to lift the ambiguity.

In our aforementioned survey, we proposed both options and asked which of these would make them feel more confident in the car's abilities.
A majority of 54.9\% answered they would feel more confident with a visual + textual/speech output. At the same time, 38.7\% opted for only the visual output.
6.3\% indicated that they would not feel more confident with any of these two options.
Hence, based on the results of the survey, in this paper we propose and evaluate a novel model that can output the source of ambiguity on the screen (visual output) or, in addition to the former, also generate a question geared towards the passenger (textual/speech output).
However, for simplicity reasons, we create a model that outputs textual questions instead of speech. \textcolor{black}{The extension from text to speech should be possible in the future with the large amount of research and improvements in this area \citep{taigman2017voiceloop, ren2019fastspeech, valle2020flowtron}}.

When passengers give a command to a self-driving car that refers to a specific object, a first step in the command understanding is to detect this object \citep{Deruyttere2020C4AV}. This task is often called \textit{\acrfull{vg}} in the literature.
However, the issued command could introduce some uncertainty for the VG model.
Hence, we propose a novel \textit{\acrfull{pipeline}}, to make the VG model capable of coping with uncertainty or ambiguity caused by the command and by the visual scene perceived by the car.
A visualisation of the complete \gls{pipeline} can be found in Figure~\ref{fig:proposed_system}.
To use \gls{pipeline}, the softmax output of the \gls{vg} model is first obtained by giving it an image, a free form natural language command, and a set of objects that have been extracted with a \acrfull{rpn}.
Next, \gls{pipeline} measures, based on the output of the \gls{vg} model, if the command can be easily interpreted (step \textbf{1} in the Figure), or whether the system is uncertain about some objects as being the referred object in the command.
To detect this uncertainty, we evaluate many different methods (e.g., ensembles, temperature scaling) and combinations of these methods.
Furthermore, we design a novel set of restrictions to limit the number of predicted uncertain objects while still maintaining a high increase in performance.
In the case of uncertainty, \gls{pipeline}
can return a \textbf{visual} output, and optionally, a \textbf{textual} output too.
We design a robust two-layer LSTM referring expression generator that can efficiently use image features for the textual output.
Furthermore, we propose a varying set of attributes that can uniquely describe the objects (e.g., color, distance, action) and experiment with using them in the referring expression generator.

The proposed \gls{pipeline} is extensively evaluated on the Talk2Car dataset~\citep{deruyttere2019talk2car}.
For the \textbf{textual} output of \gls{pipeline} we have extended the dataset with expressions that discriminatively refer to specific objects, also called \textit{referring expressions}.
Next to these expressions, we also annotate the objects with attributes to see if this helps in generating better expressions, for which we propose the \textit{\acrfull{reg_model}} %
(Step \textbf{2} in Figure~\ref{fig:proposed_system}).
The Talk2Car dataset was chosen over other datasets since it already contains many different modalities such as radar, LiDAR, instance segmentation, video, map annotations, etc., some of which are used in this work.
We separately evaluate each component of the system and assess the improvements in correctly understanding the commands of the Talk2Car dataset when based on the proposed human-machine interaction.

The contributions of this study are fourfold:
\begin{enumerate}[noitemsep,topsep=0pt]
     \item We evaluate different uncertainty detection methods and their combinations. We present a novel set of constraints valuable in a self-driving car situation that keep a balance between a low amount of uncertain objects presented to the passenger and high accuracy in solving the uncertain situation.
    \item Based on the detected uncertainty, we propose a method that generates questions given the objects' visual characteristics that cause uncertainty.
    \item We quantitatively evaluate the uncertainty detection model and the question generation approach.
    \item Finally, human judges qualitatively evaluate the validity of the generated questions in solving the uncertainty.
\end{enumerate}
In Section~\ref{sect:related_work}, we discuss the related work.
In Section~\ref{sect:full_pipeline} we describe the \gls{pipeline} by explaining the \gls{vg} model used in {\color{black}Subsection~\ref{sect:CU_model}}, how uncertainty is detected in {\color{black}Subsection~\ref{sect:detecting_hesitation}}, how objects are described in Section~\ref{sect:language_gen}, and more specifically how a question is generated in {\color{black}Subsection~\ref{sect:generating_questions}}.
Section~\ref{sect:experiments_and_discussion} discusses the experiments and the used dataset.
Finally, Section~\ref{sect:conclusion} explains the conclusion and possible future work.

\section{Related Work}\label{sect:related_work}
The main topics we discuss in our related work are object detection in terms of the command ({\color{black}Subsection~\ref{sect:rel_object_detection}}), uncertainty detection ({\color{black}Subsection~\ref{sect:rel_uncertainty_detection}}), and referring expression generation ({\color{black}Subsection~\ref{sect:rel_referring_expression}}).
\subsection{Detection of the Referred Object of the Command}\label{sect:rel_object_detection}

Several approaches for the \gls{vg} task (i.e., finding the object referred to in the command) in a self-driving car setting have been proposed. One approach implements a multi-step reasoning neural network where the network's reasoning process is guided by focusing on the natural language command sub-parts in different reasoning steps~\citep{yu2018mattnet, deng2018visual, AAAI20Deruyttere}.
Another paradigm is using a graph-neural network~\citep{liu2020learning,wang2019neighbourhood}.
Current most successful approaches use pre-trained language models to encode the language command~\citep{Lu_2020_CVPR,chen2020uniter}.
For the current study, we use the model from \citet{CMSVG} which uses a pre-trained Sentence-BERT by  \citet{reimers2019sentencebert} to encode commands, and a pre-trained EfficientNet-b2 by \citet{tan2019efficientnet} to encode objects detected in the image\footnote{We use this model as its code is available at \url{https://github.com/niveditarufus/CMSVG}}.
However, detecting the referred object in the command is not always correct, hence the importance of accurate uncertainty detection and quantification.

\subsection{Uncertainty Detection}
\label{sect:rel_uncertainty_detection}
Currently, we rely on numerous artificial intelligence systems to help us with decision making. Detecting when the system's prediction is uncertain and accurately quantifying this uncertainty are research topics of increasing interest. Especially when decisions are taken in real-world situations, uncertainty detection is of primordial importance.
The probabilistic output of a classification model, for instance, obtained with a softmax function, is not always a reliable reflection of the model's confidence.
Even with high \textit{softmax} values, a model can still be uncertain about its prediction~\citep{gal2016uncertainty}.
This is because the classifier, and specifically deep neural networks, tend to be poorly calibrated and overconfident in its predictions ~\citep{guo2017calibration}.

To overcome this issue, \textit{temperature scaling} is applied to the \textit{softmax} outputs a posteriori~\citep{guo2017calibration}.
Additionally, Bayesian Neural Networks could be used~\citep{mackay1992practical}, which detect uncertainty by specifying a prior distribution over the parameters of a neural network.
After training the network on data, a posterior distribution is used to estimate the uncertainty~\citep{lakshminarayanan2017simple}.
However, these networks are not well suited for many real problems as they are computationally complex and challenging to train ~\citep{gal2016dropout,kendall2017uncertainties,lakshminarayanan2017simple,ayhan2019expert}.
An alternative that approximates the Bayesian Neural Network's behavior is using an ensemble of networks to quantify uncertainty~\citep{lakshminarayanan2017simple}.
An ensemble of networks can also be used to re-calibrate an existing network.
Instead of training and running multiple networks,
\citet{gal2016dropout} propose to use Monte Carlo Dropout (MC Dropout)~\citep{srivastava2014dropout}
which applies the dropout during both training as well as during inference.
During inference, this method simulates having an ensemble of networks by sampling.

Quantifying uncertainty has made a successful contribution to several applications. For instance,
\citet{ayhan2019expert} have created a system to detect diabetic retinopathy that also reports its uncertainty by using data augmentation during inference.
Detecting uncertainty with MC Dropout in 3D object detection with LiDAR also resulted in improved accuracy~\citep{feng2018towards}.
To generate better depth maps when jointly using LiDAR and images, \citet{van2019sparse} use a network with a branch for local information and another for global information. In addition, each branch learns a confidence map in an unsupervised manner. Based on these confidence maps, the network knows which branch is more certain and can give it more weight.
\citet{lee2018ensemble} use a Bayesian ensemble to cope with sensor failures in an autonomous vehicle.
\citet{xiao2019quantifying} employ MC Dropout to measure uncertainty in natural language tasks and show that this improves accuracy.

In this paper, we focus on methods that detect and quantify uncertainty when the system understands a natural language command given in a certain visual context. We want the method to be agnostic of the underlying \acrfull{vg} system, hence our choice of
\textit{temperature scaling} and \textit{ensemble} methods.
An important novel goal of this paper is how to adapt uncertainty detection to the situation where a passenger interacts with his or her self-driving car and how to balance the accuracy of the uncertainty detection and the limitation of the cognitive overload for the passenger.

\subsection{Resolving Uncertainty with a Visual Human-Machine Interaction}
\citet{rupprecht2018guide} propose a method to improve image segmentation models by giving a user a first estimate of the segmentation mask. Then, the user can indicate the wrongly segmented parts of the image through speech/text. In a second pass, the segmentation model updates its prediction based on the received feedback.
Instead of giving user feedback in the form of speech/text, \citet{agustsson2019interactive} propose to allow the user to draw a corrective scribble on top of the semantic segmentation to indicate the wrongly predicted parts.
\citet{branson2014ignorant} propose a hybrid human-machine vision system for fine-grained categorization.
The goal is to predict which bird is shown in the image. The machine needs to ask questions to a human to reduce its uncertainty about the predicted bird species as quickly as possible.
The human can then click on the image to indicate specific parts of the bird or tell the system which color certain parts are.

\subsection{Resolving Uncertainty by a Natural Language Interface: Question Generation}
\label{sect:rel_referring_expression}
Handling an uncertain or ambiguous situation by posing a natural language question to the user of an AI system is generally a good idea.
For instance,~\citet{xu2019asking} generate template-based questions in a knowledge-base question answering setting, where a binary classifier first detects whether a question needs to be posed by the system. Specific parts of a question are generated relying on a Seq2Seq ~\citep{bahdanau2014neural} or a Transformer~\citep{vaswani2017attention} architecture and are then joined based on a template.
This paper will use a similar approach.
A completely different approach is to extract text spans from a context instead of generating free-form questions or expressions~\citep{qi2019answering}.

Generating natural language questions about objects that cause uncertainty is related to referring expression generation (REG) where expressions uniquely describe objects compared to other objects in the same image.
 The task has similarities with the ``dense image captioning'' task~\citep{johnson2016densecap}, which creates a non-unique caption for all perceived objects in the image.
The REG task can be defined as follows: given an image and a set of objects, a model needs to generate a \textit{unique} referring expression that describes each of the given objects.
\citet{mao2016generation} use a convolutional neural network (CNN) to encode the objects in the image and pass these features to a recurrent neural network (RNN) to generate a caption describing the objects.
They use Maximum Mutual Information (MMI) training to penalize the model if it thinks that a certain expression could also be generated from another object in the same image.
\citet{liu2017referring} first train an attribute predictor to predict the attributes of objects. Then, they pass these attributes together with the extracted CNN object features to their generation module, which uses an LSTM to generate an expression.
\citet{yu2016joint} embed object features and referring expressions in a joint semantic space used to generate expressions that discriminate between objects.
The idea of a joint semantic space is followed by \citet{tanaka2018generating}, resulting in one of the top-performing methods in referring expression generation.
The difference with the work from \citet{yu2016joint} is the use of a target-centered prior where they put a Gaussian distribution over the location of the object, the features of the target object, and the sentence context under generation. \citet{tanaka2018generating} obtain similar or better results as \citet{liu2020attribute} on RefCOCO(+,g).
In this paper, we use the proposed models by \citet{yu2016joint} and \citet{tanaka2018generating} as baselines in the task of question generation.

Graph-based solutions have recently been proposed~\citep{yao2018exploring,gu2019scene} in the frame of natural language generation. These solutions stand or fall with the quality of the generated graphs~\citep{zellers2018neural}.
This paper proposes to condition the generated question on the object's label and its attributes.
These can be reliably extracted from the visual scene through the use of attribute predictors.
Additionally, we propose a method that enforces the use of these labels in the generated question.

State-of-the-art language models make use of a Transformer architecture~\citep{vaswani2017attention}, like GPT-3~\citep{brown2020language}. However, this is not the case for any of the REG state-of-the-art models.
Furthermore, in our case, the generated questions that describe uncertain objects require to be short, and their generation should be fast in the self-driving car setting.
While the REG state-of-the-art models use a more simplistic Long Short-Term Memory (LSTM) model design, we design a robust two-layer LSTM referring expression generator that can efficiently use all features for the textual output.
Furthermore, we propose some varying sets of attributes that can uniquely describe the objects (e.g., color, distance, action) and integrate them in the referring expression generator.

\section{\acrfull{pipeline}}
\label{sect:full_pipeline}
In this section, we describe the proposed system  (Figure~\ref{fig:proposed_system}), which we name \acrfull{pipeline}.
We start this section by describing the notation used in this paper ({\color{black}Subsection~\ref{sect:notation}}).
Then, we discuss the \gls{vg} model ({\color{black}Subsection~\ref{sect:CU_model}}) that detects the object that is referred to in the command in the visual scene.
Next, we move to uncertainty detection and quantification, and identify the objects proposed by the \gls{vg} model that cause the uncertainty ({\color{black}Subsection~\ref{sect:detecting_hesitation}}).
Then, we describe the generation process of the question, which is used to ask the passenger for clarification (Section~\ref{sect:language_gen}).

\subsection{Notation Introduction}\label{sect:notation}
Following the notation standards from \citet{Goodfellow-et-al-2016}, vectors will be represented as $\vv$ and their size $d$ will be represented as $\sR^d$, and
matrices will be represented as $\mM$ and their dimensions will be represented as $\sR^{a \times b}$ where $a$ is the input dimension and $b$ the output dimension.

\subsection{Visual Grounding (VG) Model}
\label{sect:CU_model}
Before using our \gls{pipeline}, we require a \gls{vg} model that finds the  referred target object \gls{m:t-obj} by the command \gls{m:com} in a certain image \gls{m:img}.
We can create such a model by using the following statement:
\begin{equation}
\gls{m:obj} = \argmax_{\gls{m:obj} \in \gls{m:objs}} p(\gls{m:obj}=\gls{m:t-obj} \vert \gls{m:img},\gls{m:com}, \m{\theta}) \label{eq:cu_goal}
\end{equation}
with \gls{m:objs} the set of all objects in the image and $\m{\theta}$ the model parameters.
For readability, we notate the probability distribution over the set of objects as \gls{m:obj-dist}, with \gls{m:vars} the set of all inputs.
Although our model is agnostic of the underlying VG model for computing this probability distribution, in this paper we make use of the CMSVG model~\citep{CMSVG} as the \gls{vg} model, since it is one of the top-performing models on the Talk2Car dataset at the time of writing.

This model uses CenterNet~\citep{duan2019centernet}
as a \gls{rpn} to extract the set of objects \gls{m:objs} objects from image \gls{m:img}.
For each object $o \in \gls{m:objs}$, EfficientNet-b2~\citep{tan2019efficientnet} extracts its L2-normalized visual feature vector of $\sR^{1000}$. The command $c$ is encoded into a feature vector of $\sR^{1024}$ by a RoBERTa~\citep{liu2019roberta} based Sentence-BERT~\citep{reimers2019sentencebert}.
The resulting vector is then mapped to a vector of $ \sR^{1000}$ using a fully-connected layer to obtain a sentence embedding with the same dimension as the visual feature vector.
Next, the sentence embedding is combined with the visual feature vector through a dot product followed by the application of a softmax function to compute the probability distribution, as shown in Eq.~\ref{eq:cu_goal}.
For a more detailed explanation, we refer the reader to the original paper by~\citet{CMSVG}.

\subsection{Jointly Detecting Uncertainty and Uncertain Objects}%
\label{sect:detecting_hesitation}
In the first part of \gls{pipeline} (step 1 in Figure~\ref{fig:proposed_system}), the uncertainty when predicting the referred object \gls{m:obj} in Eq.~\ref{eq:cu_goal} is quantified.
Furthermore, this step jointly creates a subset $O_c \subseteq \gls{m:objs}$ containing viable candidates for the target referred object {\color{black} which can be used to limit the amount of computations needed in the next step of \gls{pipeline}}.
We examine several meta-classifiers that classify the \gls{vg} model's output as correct or uncertain solely based on its created probability distribution and also return a subset $O_c$ of candidates for the referred object.
The examined classifiers all use a variety of combinations from the four elements described below: model calibration ({\color{black}Subsection~\ref{sect:calibration}}), used output function of the \gls{vg} model ({\color{black}Subsection~\ref{sect:output_function}}), uncertainty detection method ({\color{black}Subsection~\ref{sect:uncertainty_detection_methods}}), and the number of objects in the visual scene ({\color{black}Subsection~\ref{sect:influence_number_objects}}).

\subsubsection{Calibrating the \gls{vg} Model}
\label{sect:calibration}
Interpreting the pure softmax output of a classifier as its confidence is not always the best approach, as most deep networks tend to be overconfident \citep{guo2017calibration}. However, by re-calibrating the model using an \textit{Ensemble} or a posteriori \textit{Temperature Scaling}, the softmax output can be correlated better to the confidence. We will experiment if re-calibrating with
these methods aids in the uncertainty detection compared to the original softmax output.

\paragraph{Ensembles (Ens)} have successfully been used to calibrate models, as well as for measuring the
uncertainty of the prediction
\citep{lakshminarayanan2017simple}. Therefore, we create an ensemble of $E$ randomly initialized \gls{vg} models.
To compute the calibrated probability distribution for the ensemble, we use the following equation:
\begin{equation}
\label{eq:ensemble_prob}
   p_E(\gls{m:objs} \vert \Phi, \m{\theta}) =  \frac{1}{E} \sum_{e=2}^{E} p_e(\gls{m:objs}\vert \Phi, \m{\theta})
\end{equation}
where $p_e$ is the probability output of the $e$-th \gls{vg} model, and $p_E$ is the calibrated distribution of the full ensemble.
In the experiments, we will use the notation $Ens_{E}$ where $E$ indicates the number of models in the ensemble.
In addition to calibrating a model, using an ensemble can also increase the accuracy.
Hence, we also test in our experiments how much can be gained in terms of accuracy when using an ensemble.

\paragraph{Temperature Scaling (TS)}~\citet{guo2017calibration} update the regular softmax function to include a hyper-parameter $\gls{m:tau}$:

\begin{equation}
\softmax\left(z_{i}\right)=\frac{\exp{( z_{i})}/\gls{m:tau}}{\sum_{j=1}^{N} \exp{(z_{j})}/\gls{m:tau}},
\label{eq:temperature_eq}
\end{equation}
where $\vz$ are the raw logits for some arbitrary classifier, $z_j$ is the $j$-th value in $\vz$, and $N$ the number of classes.
\cite{guo2017calibration} note that by examining Eq.~\ref{eq:temperature_eq}, one can observe that with $\gls{m:tau} \xrightarrow{} \infty$, the probability distribution approaches a uniform distribution, and with $\gls{m:tau} \xrightarrow{} 0$, the probability distribution collapses to a point mass, representing maximum uncertainty and complete certainty respectively. When $\gls{m:tau}=1$, we have the normal softmax function.
Note that this method does not influence the model's accuracy in any way, only the calibration.

\subsubsection{Influence of other output functions}\label{sect:output_function}
Here, we investigate whether we could replace the softmax function, which is used to transform the raw logits of the \gls{vg} model into a probability distribution, with a sigmoid function.
This operation does not affect the ordering of the classes or the output range (still between 0 and 1); only the final distribution is not a proper probability distribution anymore.

Additionally, we investigate if the \gls{vg} model's raw logits could be used to detect uncertainty.
This procedure needs to apply some additional constraints (i.e., limiting the raw logits' range to a pre-defined range) placed on the raw logits to allow efficient computation.
To limit the raw logits' range to [-1,1], we use the cosine similarity of the object's feature vector and the sentence encoding.

\subsubsection{Uncertainty Detection Methods}
\label{sect:uncertainty_detection_methods}
Uncertainty detection allows directly predicting whether the model was certain about its output or uncertain.
In the latter case, the method selects the objects causing the uncertainty and collects them in the candidate set $O_c \subseteq \gls{m:objs}$.
The ideal meta-classifier has a high accuracy for detecting when the object with the highest probability proposed by the \gls{vg} model is also the correct referred object.
This requirement is vital as every output classified as certain by our meta-classifier will be unrecoverable in our system.

Furthermore, the candidate set should remain small since we are in a time-critical environment. Confronting a car's passenger with many uncertain objects in the visual scene to choose from or generating a natural language question that entails the description of many objects in the visual scene would lead to a cognitive overload.
Consequently, this situation would not allow a quick response by the passenger and the car's consequent timely action.
Thus the method should balance between maximally exploiting uncertainty and restraining the size of $O_c$ on which the question generation will be based.
The evaluated methods are described in the following paragraphs.

\paragraph{Softmax Addition (SA)}
relies on the softmax probability distribution adding up to $1$. The top-$k$ objects from $\gls{m:objs}$ are selected based on their probability from distribution $p(\gls{m:objs}\vert \Phi, \m{\theta})$, such that the sum of these $k$ probabilities is higher then the sum of remaining $\vert\gls{m:objs}\vert-k$ probabilities where $\vert\gls{m:objs}\vert$ is the number of objects in \gls{m:objs}.
With $k=1$, the model is classified as certain and with $k>1$ the model is uncertain with all top-$k$ objects in the candidate set $O_c$.

\paragraph{Centroid Agglomerative Hierarchical Clustering (CAHC)}
starts the clustering with every object forming its own cluster with the probability of the object
$p(\gls{m:objs}|\Phi, \m{\theta})$ representing it.
Next, the method iteratively merges pairs of clusters if their centroids' absolute distance is smaller than a parameter $\delta$.
The centroid of the merged cluster is computed as the average of the probabilities of its objects.
When no more clusters can be merged, i.e., there are no clusters within a distance $\delta$ of each other, the one with the largest probability is selected.
If there is only one object in the cluster, the model is classified as certain.
Otherwise, it is uncertain with all the objects in the highest probability cluster being members of the candidate set $O_c$.

\paragraph{Thresholding (SoftTr, SigmTr, RLT)}
makes use of a threshold $\eta$ for classifying the model as certain or not.
In case of the softmax output of the model, the threshold (trained on the validation set) is applied over the probability distribution $p(\gls{m:objs} \vert \Phi, \m{\theta})$ to create the candidate set $O_c$ as follows:
\begin{equation}
\forall \gls{m:obj} \in O_c \iff p(\gls{m:obj} \vert \Phi, \m{\theta}) > \eta \label{eq:candidate_set}
\end{equation}
In case that only a single object is part of the candidate set, the model is classified as certain. Otherwise, it is uncertain.

When the softmax output function is changed with a function from {\color{black}Subsection~\ref{sect:output_function}}, the probability distribution in Eq.~\ref{eq:candidate_set} is replaced by the new output distribution.
In the experiments we will refer to softmax thresholding as \textit{SoftTr}, sigmoid thresholding as \textit{SigmTr}, and raw logits thresholding as \textit{RLT}.

\paragraph{Jenks Natural Breaks Optimization (Jenks)}
This algorithm, which can be seen as a 1-dimensional $K$-means clustering~\citep{dent2008thematic},
tries to determine the best arrangement of values into different classes. It does this by minimizing each class's average deviation from the class mean, while maximizing each class's deviation from the other group's mean. In other words, the method seeks to reduce the variance within classes and maximize the variance between classes~\citep{jenks1967data}. We say that the model is certain if the cluster with the highest probability only contains one object and uncertain otherwise. In the latter case, the objects in this cluster with the highest probability score are the uncertain objects.
The amount of clusters $k$ is found by finding a $k$-value that minimizes class means' squared deviations.

\paragraph{Ensemble Voting (EV)}
This method is only usable with an ensemble of \gls{vg} models.
We let every \gls{vg} model in the ensemble vote for the command's referred object. Each model casts a vote for the referred object.
If all the models vote for the same object, we say the ensemble is certain and uncertain otherwise. In the latter case, the different objects that have received votes represent the uncertain objects.

\subsubsection{Influence of the number of objects in \gls{m:objs}}
\label{sect:influence_number_objects}
The fourth and last category investigates the influence of the number of objects \gls{m:objs} given to the \gls{vg} model.
Remember, we want to limit the number of objects detected as uncertain when interacting with the car's passenger.
Hence, we will investigate the influence of selecting the top-$k$ objects with the highest probability from the RPN.

Additionally, another way to reduce the number of objects in \gls{m:objs}, is to predict the class (or its superclass, for instance, ``car'' has as superclass ``vehicle'') of the object referred to by the command. We will refer to this predicted class based on the command $c$ as $r_c$.
All objects whose predicted class from the RPN is not equal to $r_c$ can then be ignored.
To investigate this, we create the following referred object class predictors that take as input the command $c$ and predicts the referred object's class $r_c$.
In the experiments we will refer to removing all the objects which do not belong to the same class as \textit{Class Filtering (CF)} and removing all objects which do not belong to the same superclass as \textit{Superclass Filtering (SCF)}.

\paragraph{Bidirectional-LSTM (Bi-LSTM)}
Our first $r_c$ predictor is a Bi-LSTM that takes the full command as input and embeds each word as a vector of $\sR^{512}$.
Afterward, an embedding of $\sR^{512}$ is created to represent the command using the Bi-LSTM.
Finally, this sentence embedding is then passed to a linear layer to predict the referred object class.

\paragraph{Bi-LSTM with attention (Bi-LSTM Att.)}
Some words are more important than others to know which object is being referred to by the command.
To exploit this, we first encode the $n$ word embeddings of size $\sR^{512}$ from the command into an embedding $\vs \in \sR^{512}$ with a Bi-LSTM.
Then, we use the following equation to compute a softmax distribution over the word embeddings of the command:

\begin{equation}
\va = softmax(f(\vs \bullet \mW))\label{eq:bilstm_att}
\end{equation}
we define $\bullet$ as the operator that copies the first operand ($\vs$)  to the dimensions of the second operand ($\mW$)  and then applies the hadamard product, $\mW \in \sR^{n \times 512}$ represents the command's word embeddings, and $f$ is a linear layer of $\sR^{512\times1}$.
The attention vector $\va \in \sR^{n}$ is used in the following manner to produce a sum weighted sentence embedding of $\sR^{512}$:

\begin{equation}
    \vr = \sum_{i=0}^{n-1} a_i \v{W_{i}}\label{eq:attsum}
\end{equation}
where $\v{W_i}$ is the $i$-th word embedding. The resulting vector $\vr$ is passed through a linear layer to predict $r_c$. The idea behind the vector $\vr$ is that the words that contain the most information for the classification, will have a higher softmax score than other words and thus have a higher weight in the sum.

\paragraph{Sentence-BERT}~\citep{reimers2019sentencebert} is the final model that will be explored to predict $r_c$. Sentence-BERT is a BERT model~\citep{devlin2018bert} trained as a siamese neural network with tied weights. The goal of the model is to create semantically meaningful sentence representations.
For more information about this model, we refer the reader to the original paper.

\section{Describing the Objects that Cause the Uncertainty}
\label{sect:language_gen}
The next step in \gls{pipeline} (part 2 in Figure~\ref{fig:proposed_system}) is displaying the uncertain objects to the passenger. Yet, optionally, a textual question can also be provided.
This question can be created by generating referring expressions for uncertain objects and chaining them together.
{\color{black} As we are only interested in describing the uncertain objects, we can leave out the other objects and thus save computations.}

In this section our proposed referring expression model \acrfull{reg_model} is described.
Since the model makes use of attributes, we first describe the attribute predictor in {\color{black}Subsection~\ref{sect:attribute_predictor}}. Then we will define \gls{reg_model} and possible variations in {\color{black}Subsection~\ref{sect:expression_gen}}.
We finish this section with the baselines with which our models are compared.

\subsection{Attribute Predictor}
\label{sect:attribute_predictor}
Our goal in this section is to generate qualitative referring expressions. Therefore, we investigate whether better expressions can be generated when using attributes of the objects extracted from an attribute predictor.
We identify three types of attributes that are interesting for the task at hand: the \textit{location} of the object relative to the car, the \textit{color} of the object, and the \textit{action} performed by the object.
We experiment with predictors that predict each of these types separately and a predictor that jointly predicts action and color.

For predicting the object's action, we first create an embedding of $\sR^{512}$ for the (predicted) class of the object. Then, we pass the image and its cropped object, after resizing both to $224 \times 224$, through ResNet-152~\citep{he2016deep} pre-trained on ImageNet~\citep{deng2009imagenet}, with the last linear layer removed resulting in a vector of $\sR^{2048}$ for both the object and the image.
We concatenate the object, image, and class embedding into a vector of $\sR^{4608}$.
This is then passed through a first linear layer of $\sR^{4608 \times 1024}$ and a second linear layer of $\sR^{1024 \times 11}$, where $11$ is the number of actions.
\newline
For predicting the object's color, a similar network is used as the one for predicting the action.
Now, only the object is cropped and passed through a pre-trained ResNet-152, resulting in a vector of $\sR^{2048}$.
This is passed through two linear layers of $\sR^{2048 \times 1024}$ and $\sR^{1024 \times 12}$, respectively, where $12$ is the number of possible colors.
\newline
For jointly predicting both action and color, we use the two models described above in one network by using the same ResNet-152 to extract region features and by having two output heads on top of the ResNet, one for the colors and one for the actions.

For predicting the location of the object, we experiment with six model options. The first is (1) a two-layer neural network with matrices $\mathbb{R}^{3 \times 100}$ and $\mathbb{R}^{100 \times 3}$.
This network takes the vector %
$[x/imgW, (x+w/2)/imgW, (x+w)/imgW]$
as input where $x$ and $w$ are the bottom right coordinate and the width of the object respectively and $imgW$ is the width of the image.
The remaining options are (2) a Decision Tree, (3) a Random Forest with ten estimators, (4) a Support Vector Machine, (5) a Support Vector Machine with RBF kernel, (6) and Logistic Regression. %
For the experiments, see {\color{black}Subsection~\ref{sect:result_attr_pred}}.

\subsection{Generating Referring Expressions} \label{sect:expression_gen}
\begin{figure*}[t]
\centering
\begin{subfigure}[t]{0.473\textwidth}
     \centering
     \includegraphics[width=\textwidth]{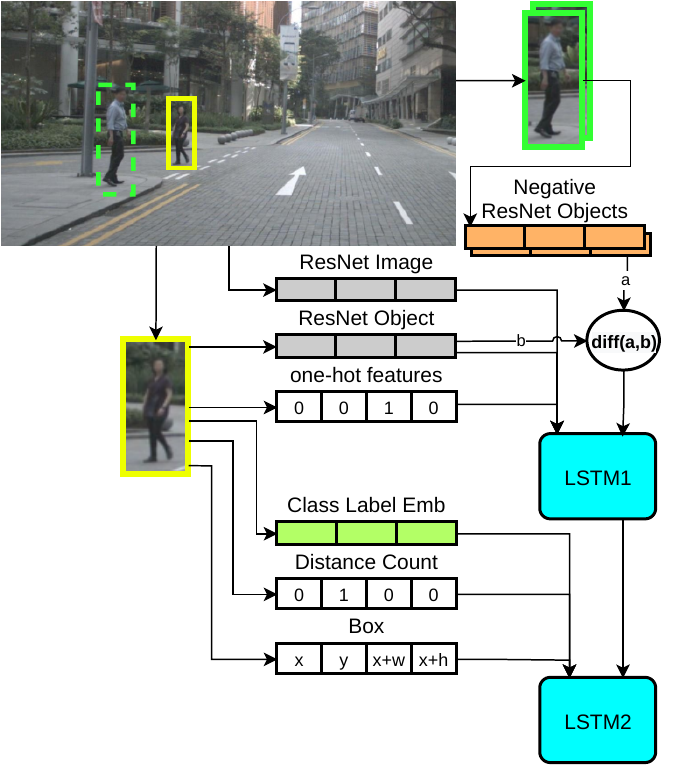}
     \caption{\gls{reg_model}-hot}
     \label{fig:a_reg_hot}
\end{subfigure}
\hfill
\begin{subfigure}[t]{0.48\textwidth}
     \centering
     \includegraphics[width=\textwidth]{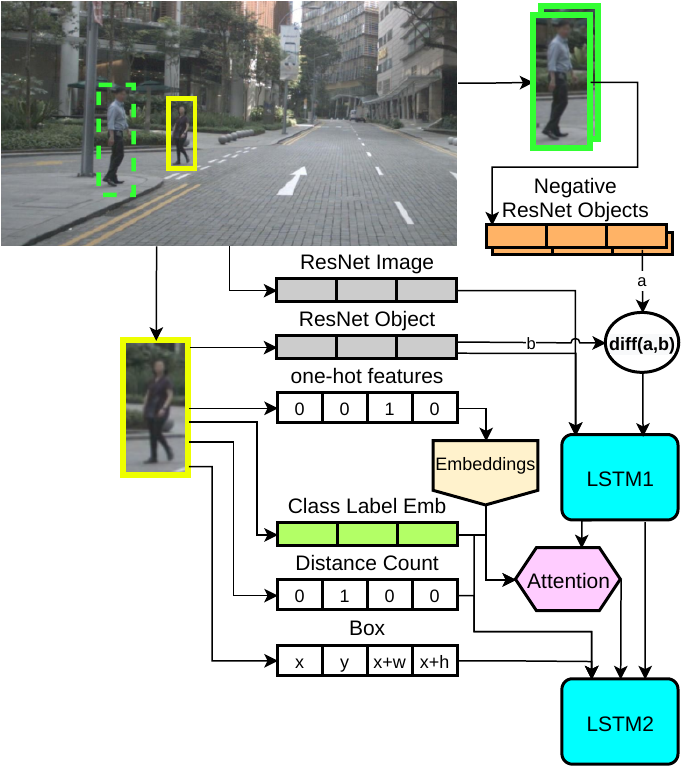}
     \caption{\gls{reg_model}-att}
     \label{fig:a_reg_att}
     \end{subfigure}
\caption{Flowcharts for the \acrshort{reg_model} variations.}{The two attribute variations in the \acrfull{reg_model}. On the left is the \gls{reg_model}-hot, where the one hot features for the different attributes (action, color, and location) are the inputs to the first LSTM layer. On the right is the \gls{reg_model}-att, where the embedding is extracted for each attribute. These embeddings are then processed through an attention layer guided by the hidden state of the first LSTM layer. The attention's output is then passed to the second LSTM layer. Best viewed in color.
}
\label{fig:lang_gen_model}
\end{figure*}

The referring expression model's goal is to describe objects in such a way that they are uniquely distinguishable from other objects.
Therefore, we hypothesize that knowledge regarding the attributes of an object can contribute to the generation of the expressions.
Hence, we experiment with several ways of providing the attributes of an object, the distance count\footnote{Counted based on LiDAR data and considering only objects of the same type for counting.
We assume that the LiDAR data is freely accessible to generate this count.}, and the class label to our \acrfull{reg_model}.
The different variations of \gls{reg_model} are visualised in Figure~\ref{fig:lang_gen_model}.

We start by describing a basic Convolutional Neural Network and LSTM (CNN-LSTM) model in {\color{black}Subsection~\ref{sect:cnn_lstm}} that does not use attributes and will serve as a control model.
Next, we introduce the \gls{reg_model} that uses attributes and define its variations in {\color{black}Subsection~\ref{sect:a_reg}}.

\subsubsection{Basic Control Model (CNN-LSTM)}
\label{sect:cnn_lstm}
This basic CNN-LSTM model is designed
to validate the contributions of the object properties.
Therefore, a minimal set of features is provided to the model.

Since we want this model to be powerful, we take inspiration from the well known Bottom-Up Top-Down model \citep{anderson2018bottom}.
This model was originally designed for image captioning, where a set of image features is extracted for each object in the image, and a two-layer LSTM with attention is used for the caption generation.
However, we only have a singular object, thus we drop the attention mechanism.
The goal of the first layer to process global information and combine it with the context from the generated expression so far.
The goal for the second layer is to use all information received for predicting the next token in the expression, since the output of the second layer is directly passed to the fully-connected output layer.

\paragraph{Box Feature LSTM (CNN-LSTM-box)}
Given an image \gls{m:img} and an object $\gls{m:obj}\in\gls{m:objs}$, we use ResNet-152 \citep{he2016deep}, without the final prediction layer, to extract object feature vector \gls{m:f-obj-dim}.
Next, we define the following two layer LSTM setup:
\begin{align*}
\gls{m:ht1} &= LSTM_1(\gls{m:i1}, \gls{m:ht1-1})\\
\gls{m:ht2} &= LSTM_2(\gls{m:i2}, \gls{m:ht2-1})\\
\end{align*}
where $\vh_l^{(t)}$ is the hidden state of layer $l$ at timestep $t$, and $\vi_l$ is the input of the layer $l$. For the CNN-LSTM model, the inputs are defined as follows:
\begin{align}
\gls{m:i1} &= [Emb(\gls{m:expr_t-1}); \gls{m:ht2-1}; \gls{m:f-obj}] \label{eq:input_l1}\\
\gls{m:i2} &= \gls{m:ht1} \label{eq:input_l2}
\end{align}
where \gls{m:expr_t-1} is the expression token at timestep $t-1$ for which we extract the embedding using function $Emb$, and ';' is the concatenation %
operator.
We initialize the hidden states $\v{h_l}$ with $zeros$ and every sentence at $t=0$ starts with a special start-of-sentence token.

\paragraph{Adding Global Image Feature Vector (CNN-LSTM-full)}
Adding the global image feature vector \gls{m:f-img} can help the model recognize the context in which the object appears in the image.
Therefore, we extract the image feature vector \gls{m:f-img-dim} with the same ResNet-152 network as used for \gls{m:obj}. Now we replace Eq.~\ref{eq:input_l1} with Eq.~\ref{eq:input_l1_img}:
\begin{equation}
\gls{m:i1} = [Emb(\gls{m:expr_t-1}); \gls{m:ht2-1}; \gls{m:f-obj}; \gls{m:f-img}] \label{eq:input_l1_img}
\end{equation}

\paragraph{Adding Difference Features (+diff)}
To help the model distinguish between different objects, an additional average of the object features from other boxes can be provided to the model \citep{yu2016joint}. The five closest objects of the same category as the object we are trying to describe are selected from the remaining candidates $\gls{m:objs}\backslash\gls{m:obj}$, with $\gls{m:obj}$ the previously selected object.
We extract the object features for the closest boxes, subtract the original box from them, and take the average, resulting in the $\gls{m:f-obj}_d\in\sR^{2048}$ and can be added to the input of the first LSTM layer by concatenating it to the input and updating Eq.~\ref{eq:input_l1} or Eq.~\ref{eq:input_l1_img}.

\paragraph{Loss Function}
To define the loss function for our referring expression models, we use the \acrfull{mmi-mm} from \citet{mao2016generation}.
The loss consists of two parts, where the first is the cross-entropy loss often used for language generation tasks:
\begin{equation}
\Ls_{ce} = -\sum \log P(\hat{X}=X^*\vert \gls{m:img},\gls{m:obj}^{(pos)},\m{\theta})
\end{equation}
Where $\hat{X}$ is the predicted expression, $X^*$ the target ground-truth expression, \gls{m:img} an image with $\gls{m:obj}^{(pos)}$ one of the candidate objects that forms a positive pair with the target expression, and $\m{\theta}$ are the model parameters.

To make sure that the expressions generated for each of the objects in a single image are unique,
a Maximum Mutual Information (MMI) constraint is added to the loss:
\begin{multline}
\Ls_{MMI} = \sum
max(0,M - \log P(\gls{m:expr}=\gls{m:target_expr}\vert \gls{m:img},\gls{m:obj}^{(pos)},\m{\theta}) ~+ \\
\log P(\gls{m:expr}=\gls{m:target_expr}\vert \gls{m:img},\gls{m:obj}^{(neg)},\m{\theta}))
\label{eq:loss_mmi}
\end{multline}

Where the second term enforces a minimum difference of margin $M$ between the expression generated for the positive candidate object $\gls{m:obj}^{(pos)}$ and some negative candidate object $\gls{m:obj}^{(neg)}$.
The final loss is given by:

\begin{equation}
\Ls = \Ls_{ce} + \gls{m:weight_mmi} \Ls_{MMI}
\label{eq:full_mmi_mm_loss}
\end{equation}

The second term is multiplied with weight \gls{m:weight_mmi} between zero and one.
The negative object $\gls{m:obj}^{(pos)}$ is randomly selected from the entire set of object candidates without the positive object $\gls{m:objs}\backslash\gls{m:obj}^{(pos)}$.

\subsubsection{\acrlong{reg_model} and its Variations (\acrshort{reg_model})}
\label{sect:a_reg}
The simplistic CNN-LSTM model only uses the features directly obtained from the image pixels.
In this section, we propose the \acrfull{reg_model} that can efficiently use several extracted properties of the object:
\begin{itemize}[noitemsep,topsep=0pt]
    \item The bounding box location property vector \gls{m:f-box}.
    \item The distance count property vector \gls{m:f-cnt}.
    \item The attribute properties vectors for actions \gls{m:f-act}, colors \gls{m:f-col}, and the locations \gls{m:f-loc}.
    \item The class label embedding vector \gls{m:f-cls}.
\end{itemize}

To reiterate, the CNN-LSTM consists of two LSTM layers, with the first layer's goal to process global information and combine it with the context from the generated expression so far. The second layer's goal is to use all information received for predicting the next token in the expression.

Since the object is only a part of the image, it can be useful for the model to know where in the image the box is located, as well as the size of the box. Therefore, we construct the bounding box vector as follows \begin{equation}
    \gls{m:f-box} = [\frac{x}{W}; \frac{y}{H};\frac{x+w}{W};\frac{y+h}{H}]
\end{equation}
where $x$ and $y$ are the coordinates of the top left corner, $w$ and $h$ are the width and height of the box in pixels, and $W$ and $H$ are the width and height of the image.

For the distance count, we make use of the LiDAR data available in the Talk2Car dataset.
This provides us with a point map defining the distance for every point.
Based on several experiments, we decided to take the closest LiDAR point that lies in a bounding box, to represent the distance of that bounding box.
We sort the bounding boxes that belong to the same class, the same spatial location (left, right, in front), and the same color, based on distance, and assign integer labels for the object's count.
We map the count to a one-hot vector \gls{m:f-cnt-dim}, where every count higher than five is mapped to the final index.

Since these properties are useful for expression generation, we add them as input to the second layer of the LSTM model, so we update Eq.~\ref{eq:input_l2} as follows:
\begin{align}
\gls{m:i2} &= [\gls{m:ht1}; \gls{m:f-box}; \gls{m:f-cnt}] \label{eq:input_l2_prop}
\end{align}

There are several options for integrating object label and attribute label information. First, we define the feature vectors; next, we describe how we apply them in the models: \gls{reg_model}-Att, \gls{reg_model}-Hot, and \gls{reg_model}-Full.

\paragraph{Attribute and Class Labels}
The methods for extracting the attribute labels are discussed in {\color{black}Subsection~\ref{sect:attribute_predictor}}.
These provide us with a probability distribution for the actions \gls{m:f-act-dim}, the colors \gls{m:f-col-dim}, and the locations \gls{m:f-loc-dim}.
While one could use these probability distributions as input we found that converting them to one-hot-encodings, for each attribute type, yields the best result. The conversion is done such that:

\begin{equation}
\hat{\vx} =
\begin{cases}
    1   & \quad \text{if } \argmax(\vx)\\
    0   & \quad \text{otherwise }
  \end{cases}\label{eq:2onehot}
\end{equation}
where $\vx$ is some probability distribution vector. We further evaluate the integration of attributes in the question generation in two settings, using the one-hot encodings directly or making use of an attention mechanism over the attributes.

Similarly, we obtain the class label. In {\color{black}Subsection~\ref{sect:CU_model}} we describe how we use the CenterNet to predict the probability distribution over classes \gls{m:f-cls-dim}, which we again convert to one-hot-encodings using Eq.~\ref{eq:2onehot}.

\paragraph{\acrshort{reg_model}-Att}
The first option for integrating the attribute and class labels in the model is by using an attention mechanism, so the model can decide which of these features is important at what time step in the sequence. Therefore, we collect the Glove embeddings \citep{pennington2014glove}, for each of the attribute types, such that
\begin{equation}
\gls{m:m-attr} = [Emb(\gls{m:f-act}), Emb(\gls{m:f-loc}), Emb(\gls{m:f-col}), Emb(\gls{m:f-cls})]
\end{equation}
where \gls{m:m-attr} is the matrix with each row an embedding feature vector.
Now we implement an attention mechanism identical to the one described in Equations~\ref{eq:bilstm_att} and~\ref{eq:attsum}. However, we replace $\mW$ with the attribute embeddings \gls{m:m-attr}, and $\vs$ is replaced with  \gls{m:ht1}, so that different embeddings are selected at every timestep. We shall rename the resulting output to \gls{m:f-attr}.

Now we simply concatenate this to the input vector from Eq~\ref{eq:input_l2} so we get:
\begin{align}
\gls{m:i2} &= [\gls{m:ht1}; \gls{m:f-box}; \gls{m:f-cnt}; \gls{m:f-attr}] \label{eq:input_l2_prop2}
\end{align}

\paragraph{\gls{reg_model}-Hot}
Instead of directly inserting the attribute and class labels, we could also provide these as global information. Therefore, we design a variation of the model in which the one-hot-encodings are entered as input in the first LSTM layer, by updating Eq.~\ref{eq:input_l1_img} to the following:
\begin{align}
\gls{m:i1} &= [Emb(\gls{m:expr_t-1}); \gls{m:ht2-1}; \gls{m:f-obj}; \gls{m:f-img}; \gls{m:f-act}; \gls{m:f-loc}; \gls{m:f-col}; \gls{m:f-cls}] \label{eq:input_l1_hot}
\end{align}

\paragraph{\gls{reg_model}-Full}
The final option for the \gls{reg_model} is to use the full set of features. This is a combination of both the attention mechanism from \gls{reg_model}-att as well as the one-hot-encodings from \gls{reg_model}-hot, as discussed in the last two paragraphs.
Both the updated equations for the input into layer one from Eq.~\ref{eq:input_l1_hot} and the updated input into layer two from Eq.~\ref{eq:input_l2_prop2} are used.

\paragraph{Optional Class Guidance (+Cls)}
Because it is important that a generated expression clearly refers to a certain object, it could be beneficial to always provide the model with knowledge regarding the class label. Therefore, we add an option for adding the class label to the inputs for the second LSTM layer (in all equations~\ref{eq:input_l1},~\ref{eq:input_l2_prop},~and~\ref{eq:input_l2_prop2}), similar to the following equation:
\begin{align}
\gls{m:i2} &= [\gls{m:ht1}; Emb(\gls{m:f-cls})] \label{eq:input_l2_cls}
\end{align}

\paragraph{Switch for Forcing Attributes (+Switch)}
The generator must make expressions with clearly defined attributes, such that expressions become unique for the specific object. We hypothesize that guiding the model in generating the attribute words can improve the quality of the expressions. Therefore we train a special gate that allows the final predictor to \textbf{switch} between predicting words from the entire vocabulary or only from the predicted attributes and class labels for the object.

The switch that decides which predictor to use takes the hidden state from the second LSTM layer \gls{m:ht2}:
\begin{equation}
\gls{m:switch} = \sigmoid(f(\gls{m:ht2}))
\end{equation}
Where we use a fully connected layer as function $f$, and $\sigmoid$ is the sigmoid function. During training we use \gls{m:switch} in a switch-loss (discussed in next paragraph), but decide whether to predict an attribute based on the ground-truth expression. During inference, \gls{m:switch} is rounded such that one of the attributes is forced when $\text{round}(\gls{m:switch}) = 1$, and the regular vocabulary predictions are used when $\text{round}(\gls{m:switch}) = 0$

\paragraph{Switch Loss}
Because we do not use the output from the switch during training due to teacher forcing, we require a loss function to make sure that the switch improves over time. The switch loss can be formulated as a regression target, such that it gets close to either one or zero.

\begin{equation}
\Ls_{switch} = ~\sum_{t=1}^{T}
(\gls{m:switch}-\gls{m:target_switch})^2
\label{eq:loss_switch}
\end{equation}
where \gls{m:target_switch} is the ground-truth value for the switch and is defined as:

\begin{equation}
\gls{m:target_switch} =
\begin{cases}
    1   & \quad \text{if}~X^{(t)}~\text{belongs to the attributes vocabulary} \\
    0   & \quad \text{otherwise}
  \end{cases}\label{eq:switch2onehot}
\end{equation}

where $X^{(t)}$ is $t$-th word from the ground-truth expression $X$.

Finally, if a model uses the switch loss, the full loss is given by:

\begin{equation}
    \Ls = \Ls_{ce} + \gls{m:weight_mmi} \Ls_{MMI} + \gls{m:weight_switch} \Ls_{switch}
\end{equation}

\subsubsection{State-of-the-Art Baselines}
In this section, we discuss the baselines that are used to compare the \gls{reg_model} model described above. Our criterion for choosing a specific model is for having its code available.

\paragraph{SLR} consists of three sub-networks:
A \textit{Speaker} that follows a CNN-LSTM architecture for generating referring expressions for the target object,
a joint-embedding model, called \textit{Listener}, trained to minimize the distance between a paired object and expression representations, and a \textit{Reinforcer} that guides the model to sample more discriminative expressions.
This model uses three loss functions: a generation loss for the expression, such as the one used in this paper, an embedding loss, and a reward loss.
For a more detailed explanation, we refer the reader to the work by~\citet{yu2016joint}.\footnote{implementation available at \url{https://github.com/lichengunc/speaker_listener_reinforcer} and \url{https://github.com/mikittt/re-SLR}. We used the latter as they changed SLR to have a ResNet-152 like SR. As our model equally uses ResNet-152, we felt that it was a more fair comparison if all models used the same backbone.}

\paragraph{SR} This model is relatively similar to the SLR model, except that it consists of two sub-networks.
The first sub-network is the \textit{Speaker} that follows a CNN-LSTM architecture for generating referring expressions for the target object indicated by a Gaussian distribution over the image.
The difference with the \textit{Speaker} in SLR is that this \textit{Speaker} uses a target-centered prior, the features of the target object, and the sentence context under generation.
The second sub-network is the \textit{Reinforcer}, which works in the same way as in SLR.
This model uses two losses: a generation loss for the expression, such as the one used in this paper, and a reward loss.
For a more detailed explanation, we refer the reader to~\citet{tanaka2018generating}\footnote{Implementation available at \url{https://github.com/mikittt/easy-to-understand-REG}}.

\subsection{Generating a Question}
\label{sect:generating_questions}
Given the resulting expressions that describe each uncertain object, a possible way to generate the question for the passenger is by using the following template pattern:

\begin{table}[hpt!]
    \centering
    \begin{tabular}{cc}
        \textbf{Question :=} & \textbf{Do you mean [expr $o^1$] or} \\
        & \textbf{[expr $o^2$], ... ,or [expr $o^n$]}
    \end{tabular}
\end{table}

Where \textbf{[expr $o^n$]} is a referring expression that describes the uncertain object $o^n \in O_c$, generated by a model from Section~\ref{sect:language_gen}.

\section{Experiments and Discussion}
\label{sect:experiments_and_discussion}
As mentioned in Section~\ref{sect:introduction}, we first want to detect which objects are causing uncertainty with regard to the referred object in the command for the \gls{vg} model and then present these to the passenger.
However, we are also interested in evaluating if we can describe these uncertain objects in a question geared towards the passenger.
This question aims to give the passenger confidence in the car's abilities and provides insight into why the objects are causing uncertainty.

We start this section with some statistics of the used Talk2Car dataset.
Afterward, we discuss how we have augmented the dataset with referring expressions as these are not included in the original dataset.
Then, we report on the experiments and results for uncertainty detection and quantification, followed by experiments and results for the expression generation.
We finish this section with the final accuracy of our \gls{pipeline} and human evaluation results.
For the reader's convenience, we have grouped experiments, results and discussion with regard to the specific steps of the \gls{pipeline}.

\subsection{Talk2Car Dataset}
\label{sect:dataset}
\paragraph{Talk2Car Dataset Statistics} The Talk2Car dataset~\citep{deruyttere2019talk2car} contains images of a self-driving car driving around in Boston and Singapore (\textit{right} vs. \textit{left} hand traffic), natural language commands referring to objects, bounding boxes for 23 data classes, and the bounding box of the referred object in the command.
These images, provided by nuScenes~\citep{nuScenes}, are taken in different weather (rain or sun) and time conditions (night or day).
In total, the Talk2Car dataset provides \textcolor{black}{11,959} commands for \textcolor{black}{9,217} images.
The dataset is split into a training, validation and test set that respectively contain 8,349 (69.8\%), 1,163 (9.721\%) and 2,447 (20.4\%) commands.
From the latter, \textit{four} challenging subsets have been created. These are used to evaluate the proposed models for uncertainty detection and ambiguity resolution, their handling of different command lengths, and how they are coping with referred objects that are far away.
The ambiguity resolution in this dataset is defined as how well a model can cope with having multiple instances of the referred object class in the same image.
On average, the command contains eleven words, while 21\% of the commands' words are nouns, 21\% verbs, and around 6\% are adjectives.
In each image, we can, on average, find eleven objects and more than four objects of the same class as the referred object.
Since Talk2Car is built on nuScenes, it also provides a whole range of other modalities such as radar, LiDAR, and instance segmentation. For more information we refer the reader to \url{https://www.nuscenes.org/} and \url{https://talk2car.github.io/}.
An example of the Talk2Car dataset can be seen in Figure~\ref{fig:t2c-example}.

\subsubsection{Augmenting Talk2Car With Ground-Truth Referring Expressions and Attributes}

When training our models for generating referring expressions we have experimented with  other existing datasets that contain descriptions/referring expressions for objects such as RefCOCO~\citep{yu2016modeling}, RefCOCO+~\citep{yu2016modeling}, RefCOCOg~\citep{mao2016generation}, and CityScapes-Ref~\citep{vasudevan2018object}.
However, when applying a trained model from these datasets on Talk2Car, we saw that the domain shift was too large and resulted in unusable results.
To this end, the objects in the Talk2Car dataset are annotated (using Amazon Mechanical Turk) both with discriminative referring expressions and object attributes.
We first discuss this annotation procedure,
afterward, we analyze the resulting dataset, which we name \textit{Talk2Car-Expr}. \footnote{https://github.com/ThierryDeruyttere/Talk2Car-Expr}

\paragraph{Annotation Tool}
We developed the used annotation tool with EasyTurk~\citep{krishna2019easyturk}.
The annotation interface includes a short instruction summary explaining the annotation task's goal and a button that leads the annotators to a video explanation of the task, and a detailed textual explanation.
We ask each annotator to enter a discriminative expression and to select the attributes for the indicated bounding box of an object.
The three types of attributes are color, the action the object is performing, and where the object is (left of us, in front, right of us).

\paragraph{Analysis of the Annotations}
The dataset consists of \textcolor{black}{11,959} referring expressions, one for each object with a command in the Talk2Car dataset.
Each expression has, on average, 6.9 words, a minimum of 4, and max 22.
Each object also has three attributes: color, action, and location relative to the ego car.
The training, validation, and test splits of the dataset follow the official splits of the Talk2Car dataset.
In the appendix, the reader can find figure~\ref{fig:t2c_expr_data_statistics} where we display some statistics of the dataset.

\subsection{Evaluation Metrics}
\subsubsection{Evaluation Metric for Visual Grounding}
The Intersection over Union (IoU) between the bounding boxes of the predicted and the ground-truth object is used to measure the accuracy in the Visual Grounding task.
If this IoU $ > 0.5$, we say that the predicted object is correct.
We refer to this as $IoU_{.5}$. The IoU is defined as follows:
\begin{equation}
IoU = \frac{\text{Area of Overlap of the two boxes}}{\text{Area of Union of the two boxes}}.
\end{equation}

\subsubsection{Evaluation Metric for Jointly Detecting Uncertainty and Uncertain Objects}

To compare our methods for jointly detecting uncertainty and uncertain objects, we use the following measures and include ($\uparrow$) or ($\downarrow$) to indicate which direction is better:

\begin{itemize}[noitemsep,topsep=0pt]
    \item  $CertIoU_{.5}$ ($\uparrow$): The $IoU_{.5}$ of the used meta-classifier on the Talk2Car validation set when solely using the certain predictions.
    This is the lower bound accuracy of the meta-classifier.
    \item $CertAcc$ ($\uparrow$): The accuracy of the meta-classifier for classifying the output of the \gls{vg} model as certain. It is measured by evaluating how many times the certain outputs contain the correct answer.
    \item $CorrUnc$ ($\uparrow$): The percentage of uncertain outputs where the correct object is amongst the uncertain objects.
    \item $Th.IoU_{.5}$ ($\uparrow$): For this task, we would also like to know what the theoretical accuracy on the Talk2Car validation set for a certain meta-classifier can be.
    Remember that when outputs have been classified as uncertain and when the correct referred object is in the set of uncertain objects, it can be recovered by interaction with the passenger.
    This measure is computed as $Th.IoU_{.5} = CertIoU_{.5} + (\text{total uncertain outputs} * CorrUnc)$.
    \item $AvgUncObj$ ($\downarrow$): The average number of objects that cause uncertainty. We want this number to be as low as possible since the selected objects will be shown to the user and optionally also used to generate a question. Since giving a command to a self-driving car is a time critical task, one would like to keep the interaction with the passenger and the generated question as short as possible not to waste time.
    \item $MaxUncObj$ ($\downarrow$): The maximum amount of objects that caused uncertainty.
\end{itemize}

\subsubsection{Evaluation Metrics of the Generated Expression}
\label{sect:language_metrics}
For evaluating the generated expressions, we will use the following three metrics.

\paragraph{METEOR ($\uparrow$)}
The first metric is the METEOR score \citep{denkowski2014meteor}.
It is based on the harmonic mean of unigram precision and recall.
This method looks at all possible unigram matches between the candidate and reference sentence. Therefore, it also applies methods such as stemming and synonymy matching.
METEOR works by first creating an alignment (i.e., a mapping of unigrams) between the candidate sentence and the reference sentence.
It then computes precision and recall between the unigrams to finally compute a harmonic mean.
To account for gaps and differences in word order, a fragmentation penalty is also computed by first counting the number of n-grams that are in contiguous and identical order in both expressions.
Then, we divide this number by the total number of matched words.
The more n-grams are not adjacent, the higher the penalty.

\paragraph{ROUGE-l ($\uparrow$)}
The second metric is ROUGE-l \citep{lin2004rouge}.
This metric focuses on the longest common subsequence (LCS) between the candidate sentence and the reference sentence.
The ROUGE-l metric first computes how many words the LCS from the referenced sentence and the candidate sentence have in common compared to the length of the referred sentence and candidate sentence.
Then it uses an F-measure to get the final ROUGE-l score.

\paragraph{BLEU ($\uparrow$)}
The third metric is BLEU \citep{papineni2002bleu} which uses a modified form of precision to compare the candidate sentence to the reference sentence.
The reason to use a modified form of precision is that a model can achieve a score of 1 if it predicts the same word as many times as the length of the reference sentence.
To combat this, BLEU considers the maximum amount that a specific n-gram is present in the reference sentence and clips the count of each n-gram in the generated sentence to the maximum of this n-gram in the reference sentence.
Additionally, BLEU also adds a penalty for brief expressions.
The used value of $n$ is indicated as BLEU-n.
In this paper, we use BLEU-4.

\subsubsection{Human Evaluation Metrics}
In the human evaluation, we use the following metrics to determine the quality of the referring expression generators as part of \gls{pipeline}:
\begin{itemize}[noitemsep,topsep=0pt]
    \item For \textit{human agreement} we measure the Krippendorf's Alpha \citep{krippendorff2011computing} since it allows for any number of annotators, any number of categories or measures, and for missing data. In addition, there is no minimum sample size restriction, allowing it to be very applicable in our setting with a possible different number of objects per image.
    \item The \textit{accuracy (Acc, $\uparrow$)}, %
    measures how often the annotator correctly assigns a generated expression with its paired object.
    \item With \textit{SingleExpr ($\downarrow$)} we indicate the fraction of images where the used REG model generates the same sentence for all uncertain objects.
    \item We also use the \textit{F1} measure to compare the referring expression generators. It is taken over the assignment by annotators of generated expressions with their paired objects.
\end{itemize}

\subsection{Uncertainty - Influence of Number of Objects in \gls{m:objs} for the \gls{vg} model}
\label{sect:incluence_of_number_of_objects_exp}
As described in {\color{black}Subsection~\ref{sect:influence_number_objects}}, the number of objects might influence the number of uncertain objects.
Yet, before we investigate this issue, we ask ourselves what the influence is on both $IoU_{.5}$ and inference speed when giving the \gls{vg} model a varying amount of \gls{m:num_obj} objects.
To select these \gls{m:num_obj} objects, we sort the predicted objects from CenterNet, provided by \cite{vandenhende2020baseline}, according to their softmax confidence and take the top-\gls{m:num_obj} objects. For this experiment, the  \gls{vg} model ({\color{black}Subsection \ref{sect:CU_model}}) is trained anew for each of the top-$k$ values. During training, we use SGD with a learning rate of 0.01 and Nesterov momentum of 0.9. The learning rate is divided by 10 after 4 and 8 steps respectively.
The used batch size is 8 except for top-$64$ where we could only fit 2 batches on one RTX Titan. For top-$64$, we also change the learning rate to 0.005.

\begin{table}[pt]
    \centering
    \begin{tabular}{|c|c|c|}
    \hline
        top-k & $IoU_{.5}$ ($\uparrow$) & Inference Speed (ms) \\
        \hline
        64 & 0.656 & 187.60 \\
        32 & \textbf{0.686} & 164.44\\
        16 & 0.661 & \textbf{139.49} \\
        8 & 0.622 & \textbf{139.49}\\
        \hline
    \end{tabular}
    \caption{Results with different top-$k$ values.}{The $IoU_{.5}$ scores on the Talk2Car \textbf{test} of the used \gls{vg}-model with different top-\gls{m:num_obj} scoring objects as input. Inference speed is computed on the Talk2Car test on a RTX TITAN.}
    \label{tab:influence_of_number_of_objs}
\end{table}

In Table~\ref{tab:influence_of_number_of_objs} we see that using the top-$32$ scoring objects results in the highest $IoU_{.5}$ score.
On the other hand, using only top-$8$ objects results in the lowest score.
This should come as no surprise as with $8$ predicted objects, there is a lower chance of having the correct object amongst the predicted objects.
On the other hand, we see that top-$64$ performs worse than top-$32$. We argue that when having that many boxes, some low quality boxes can be included as well, which can eventually confuse the model.

\subsection{Uncertainty - Predicting the Referred Object Class}
\label{sect:predicting_ref_object_class}
A second way to reduce the number of uncertain objects is by creating a classifier that takes as input the command \gls{m:com} and outputs the (super)class of the object referred by the command.
This way, we can ignore all predicted objects that do not belong to the predicted class.
For this experiment, we train the three models described in {\color{black}Subsection~\ref{sect:influence_number_objects}} and evaluate their predictive accuracy on the objects' classes in the Talk2Car validation set.
Training the two LSTM models with Adam \citep{kingma2014adam}, a learning rate of 0.0001, batch size 8, and weight decay $10^{-5}$ yields the best results.
Training is stopped if during ten epochs there is no improvement on the validation set.
For the Sentence-Bert model, AdamW \citep{loshchilov2017decoupled} with a learning rate of $5 * 10^{-5}$ and batch size 64 worked best.
Additionally, we also compute the inference speed of these models.

\begin{table}[pt]
    \centering
    \resizebox{\columnwidth}{!}{
    \begin{tabular}{|c|c|c|c|}
    \hline
        Model & Cls Acc. & S-Cls Acc. & Inf. Speed (ms) \\
        \hline
        Sentence-BERT & \textbf{0.932} & \textbf{0.980} & 13.87\\
        LSTM-Att & 0.920 & 0.966 & \textbf{0.95} \\
        LSTM & 0.913 & 0.957 & \textbf{0.95} \\
        \hline
    \end{tabular}
    }
    \caption{Accuracy for predicting referred object class.}{Results for the referred object (super)class predictors on the Talk2Car \textbf{validation} set. Inference time was measured on a RTX TITAN.}
    \label{tab:predicting_referred_object_class_validation}
\end{table}

Table~\ref{tab:predicting_referred_object_class_validation} shows the accuracy and the inference speed of the referred object class predictors on the \textbf{validation} set.
Due to space constraints, we abbreviate ``Class Accuracy'' to ``Cls Acc.'' and ``Superclass Accuracy'' to ``S-Cls Acc.''.
We see that Sentence-BERT performs better than an LSTM architecture for representing the command's content.
However, it only performs marginally better than LSTM-Att but at a significantly slower pace.
The LSTM-Att model shows that by letting the model attend to certain words in the command, one can improve compared to the LSTM model.
The results also indicate that predicting the superclass (e.g., vehicle) of the object instead of the class (e.g., truck) results in higher accuracy.
In the following experiments, we will use the LSTM-Att model because of its good trade-off between speed and accuracy according to its results on the validation set.

\subsection{Uncertainty - Influence of ensemble on IoU}
\begin{figure*}[t!]
    \centering
    \includegraphics[width=\linewidth]{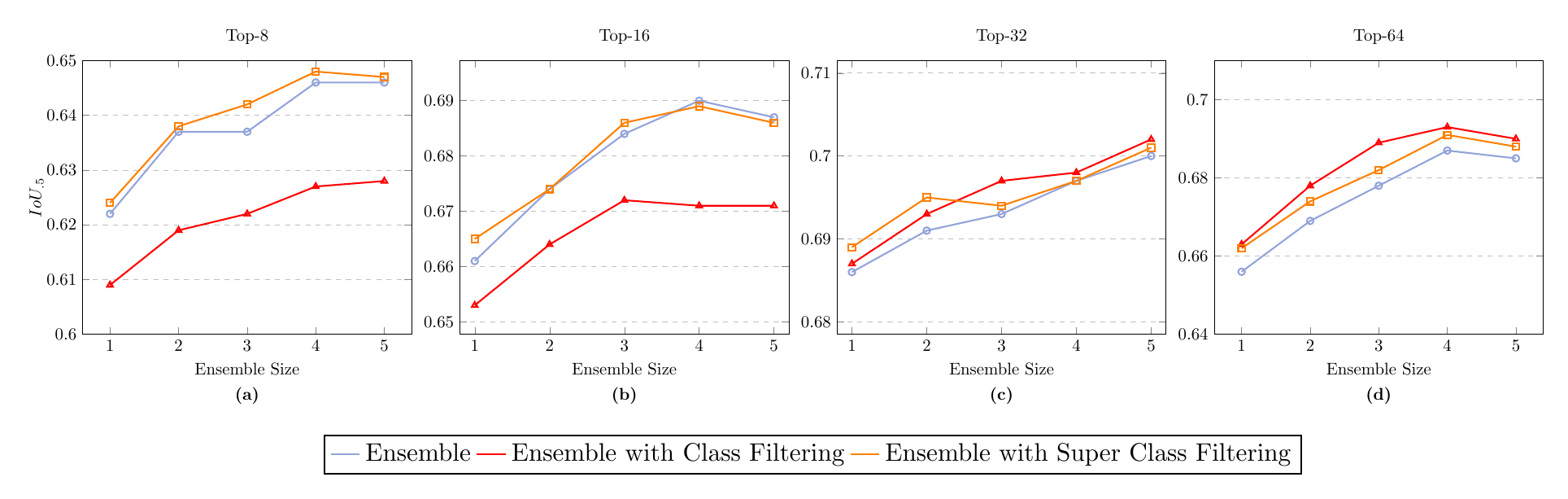}
    \caption{The influence on $Iou_{.5}$ when using an ensemble of size $E$ with different top-$k$ scoring predicted objects.}{
    At the top of each plot (a,b,c,d), we display the number of top-$k$ scoring predicted objects that have been used.
    The y-axis, shows the Intersection over Union score with threshold $0.5$ ($IoU_{.5}$).
    The x-axis, indicates the size of the ensemble ($E$).
    We also include the $IoU_{.5}$ score when $E=1$ for convenience. These values are equal to the values in Table \ref{tab:influence_of_number_of_objs}. Best viewed in color.}
    \label{fig:ensemble_accuracies}
\end{figure*}

As described in {\color{black}Subsection \ref{sect:detecting_hesitation}}, ensembles can be used to calibrate a model, but they can also influence the accuracy.
Hence in this experiment we compute the $IoU_{.5}$ on the Talk2Car test set of an ensemble ranging from $2$ to $5$ models using 8, 16, 32, or 64 top-\gls{m:num_obj} boxes.
We repeat the experiment from {\color{black}Subsection~\ref{sect:predicting_ref_object_class}}, where we first predict the (super)class of the referred object based on the command with the LSTM-Att model.
Then, we remove all objects from the top-\gls{m:num_obj} scoring predicted objects that do not belong to the predicted (super)class.
The ensemble's inference speed is equal to the inference speeds in Table \ref{tab:influence_of_number_of_objs} as these models can be parallelized.
For this experiment, we train 5 VG models for each of the top-$k$ values following the instructions in {\color{black}Subsection \ref{sect:incluence_of_number_of_objects_exp}}.
We also wish to mention that we tested the recent work by~\citet{havasi2020training} of combining multiple models of an ensemble into one model, yet we only achieved results of around $0.5 IoU_{.5}$ on Talk2Car.

The results are visualised in Figure \ref{fig:ensemble_accuracies}.
Increasing the ensemble size positively influences the $IoU_{.5}$ for all the displayed variations of the \gls{vg} model.
The top-$32$ model is consistently better than the other models in terms of $IoU_{.5}$ for any ensemble size. It also reaches an $IoU_{.5}$ of $0.70$ when the ensemble size is equal to $5$.
We also see that using Class Filtering with top-$8$ and top-$16$ reduces the performance compared to not using it.
We argue that this is because the used \gls{rpn} tends to miss-classify some of the objects with a high confidence score.
However, these miss-classified objects are still classified in the correct superclass.
For top-$32$ and top-$64$, we see that using Class Filtering and Superclass Filtering improves performance over not using it.

\subsection{Uncertainty - Jointly Detecting Uncertainty and the Uncertain Objects}
\label{sect:exp_jointly_detecting_uncertainty}
This experiment investigates which combination of the methods described in {\color{black}Subsection~\ref{sect:detecting_hesitation}} produces the best meta-classifier to jointly detect the uncertainty of a \gls{vg} model and the uncertain objects.

The following notation indicates which order the methods from {\color{black}Subsection~\ref{sect:uncertainty_detection_methods}} are used. ``CF + TS + SoftTr'' stands for first using Class Filtering to ignore the predicted objects from CenterNet, whose class is different from the predicted class obtained with LSTM-Att.
Afterward, Temperature Scaling is used on the output of the \gls{vg} model, and finally, Softmax Thresholding is used to detect if the model is uncertain and which objects are causing the uncertainty.

\begin{table*}[th]
        \small
        \begin{center}
        \scalebox{1}{
        \begin{tabular}{|l|c|c|c|c|c|c|c|c|}
        \hline
        Method & top-$k$ & $CertIoU_{.5}$ ($\uparrow$) & $CertAcc$ ($\uparrow$) &
        $CorrUnc$ ($\uparrow$) & $Th.IoU_{.5}$ ($\uparrow$) & $AvgUncObj$ ($\downarrow)$ & $MaxUncObj$ ($\downarrow)$  \\ \hline

$Ens_{5}$ + EV
& $8$ & $0.474$ &$0.822$ &$0.661$ &$0.754$ & $2.35$ & $5$ \\
$Ens_{3}$ + EV
& $16$ & $0.485$ &$0.798$ &$0.711$ &$0.764$ & $2.17$ & $3$ \\
$Ens_{4}$ + SCF + EV
& $16$ & $0.482$ &$0.800$ &$0.721$ &$0.769$ & $2.27$ & $4$ \\
$Ens_{4}$ + EV
& $16$ & $0.457$ &$0.813$ &$0.737$ &$0.780$ & $2.32$ & $4$ \\
$Ens_{5}$ + CF + EV
& $16$ & $0.530$ &$0.802$ &$0.656$ &$0.752$ & $2.24$ & $5$ \\
$Ens_{5}$ + SCF + EV
& $16$ & $0.458$ &$0.811$ &$0.733$ &$0.777$ & $2.36$ & $5$ \\
$Ens_{3}$ + CF + EV
& $64$ & $0.541$ &$0.802$ &$0.673$ &$0.760$ & $2.17$ & $3$ \\
$Ens_{5}$ + CF + EV
& $32$ & $0.521$ &$0.806$ &$0.708$ &$0.771$ & $2.34$ & $5$ \\
$Ens_{4}$ + CF + EV
& $64$ & $0.518$ &$0.828$ &$0.690$ &$0.776$ & $2.30$ & $4$ \\
$Ens_{5}$ + CF + EV
& $64$ & $0.497$ &$0.843$ &$0.700$ &$0.784$ & $2.38$ & $5$ \\
        \hline
        \end{tabular}
        }
        \end{center}
        \caption{Top-10 results of our used uncertainty detection methods.}{The top-10 results of the uncertainty detection methods from {\color{black}Subsection~\ref{sect:detecting_hesitation}} on the Talk2Car \textbf{validation} set. The measures used are explained in {\color{black}Subsection~\ref{sect:exp_jointly_detecting_uncertainty}}.}
        \label{tab:results_uncertainty_experiment_validation_set}
        \end{table*}

Because of the many possible combinations, it is unfeasible to display all of them in this paper.
Therefore, we select the combinations that have the following three restrictions: $MaxUncObj \leq 5$, $Th.IoU_{.5} > 0.75$, and $CertAcc > 0.8$, and display them in Table~\ref{tab:results_uncertainty_experiment_validation_set} in no particular order. When we allow for more than five objects, it is possible to achieve a $Th.IoU_{.5}$ up to $0.9317$. In our opinion, this is not useful because having so many uncertain objects will lead to a sensory overload for the passenger.
From the table, we see that using an ensemble jointly with class filtering and ensemble voting is an effective strategy for detecting uncertainty and uncertain objects.
Remember, next to only displaying the objects on a touch screen, we are also interested in generating questions for the passenger.
Ideally, you want to limit the number of objects that are flagged as uncertain because you want to limit the execution time of generating a question and the cognitive effort of the passenger understanding it in a self-driving car setting.
In this respect, by using Ensemble Voting, we can introduce an upper limit of the number of possible uncertain objects as this coincides with the number of models in the ensemble.
From this experiment, we decide to use $Ens_{4} + EV$ with top-$16$ objects and $Ens_{5}+CF+EV$ with top-$64$ objects
as our meta-classifier in the remainder of our experiments as both methods have high $CertAcc$ and high $Th.IoU_{.5}$.

\subsection{Visual Uncertainty Examples}
\begin{figure}[ptb]
\centering
\begin{subfigure}{0.95\linewidth}
    \centering
    \includegraphics[width=.9\linewidth]{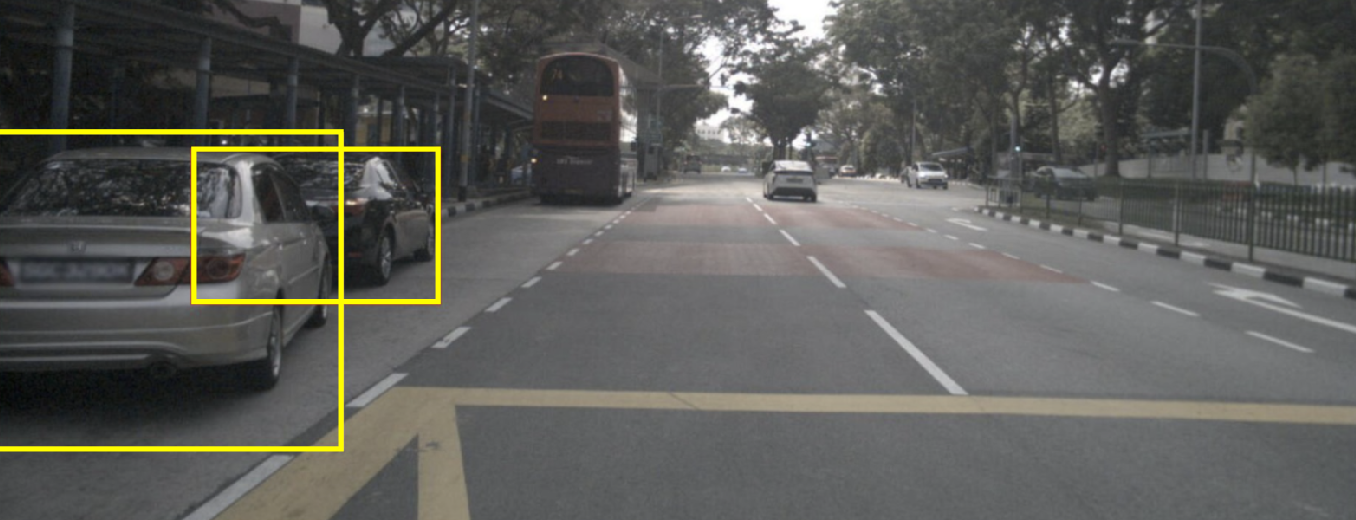}
    \caption{The objects that cause uncertainty for the command: ``Parallel park behind the car on the left''.}
    \label{fig:uncertainty_example1}
\end{subfigure}
\par\medskip
\begin{subfigure}{0.95\linewidth}
    \centering
    \includegraphics[width=.9\linewidth]{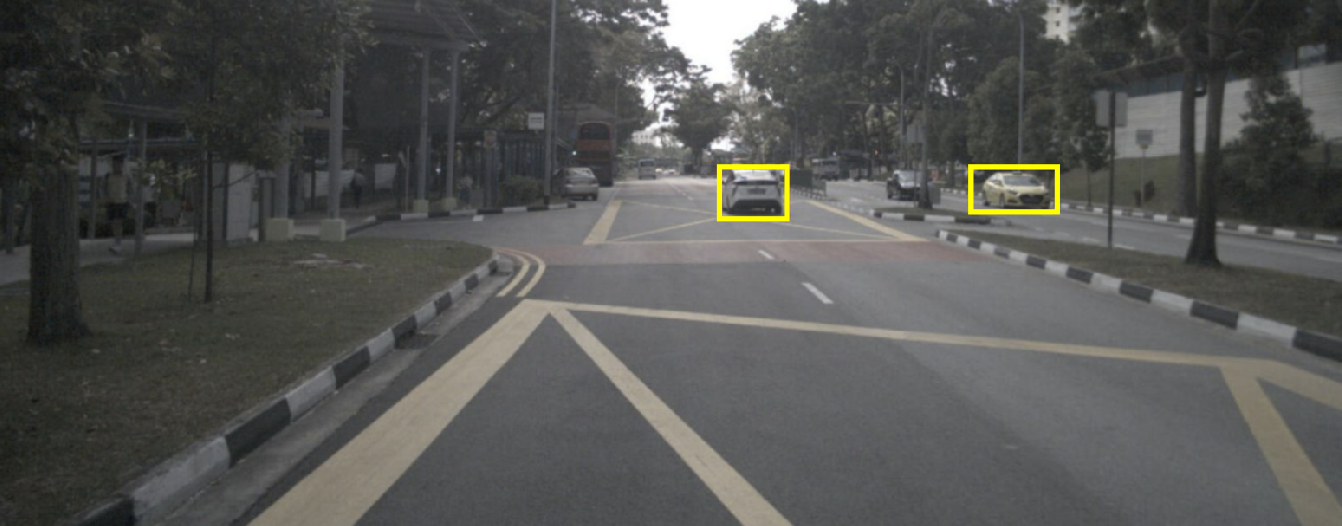}
    \caption{The objects that cause uncertainty for the command: ``Change lanes and get behind the white car''.}
    \label{fig:uncertainty_example2}
\end{subfigure}
\par\medskip
\begin{subfigure}{0.95\linewidth}
    \centering
    \includegraphics[width=.9\linewidth]{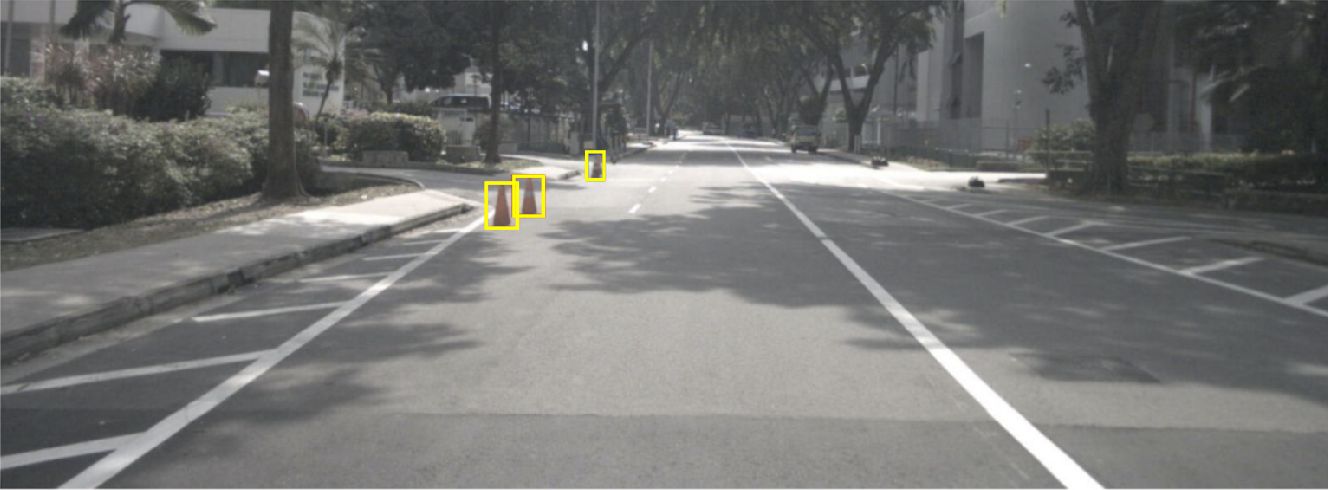}
    \caption{The objects that cause uncertainty for the command: ``After that signaling cone, turn left''.}
    \label{fig:uncertainty_example3}
\end{subfigure}
\caption{Uncertainty Examples}{Examples of uncertain objects detected in different scenes. We see that the objects flagged as uncertain by \gls{pipeline} are often from the same (super)class. Best viewed in color.}
\label{fig:uncertain_examples}
\end{figure}

We show some examples of the possible uncertain objects detected by the uncertainty models in Figure~\ref{fig:uncertain_examples}.
In the examples, there are multiple items that can cause this confusion. Often, this is between objects of the same class, which is especially clear in Figures~\ref{fig:uncertainty_example1} and \ref{fig:uncertainty_example3}. In these examples, the car has to make an action surrounding an object. In the former, the uncertainty can be resolved by the car, to ask concerning the color of the car. However, this is not possible for the cones in  Figure~\ref{fig:uncertainty_example3}. Here, the car should ask regarding the distance of the cone: \textit{Do you mean the First, second or last cone}?
In the example in Figure~\ref{fig:uncertainty_example2}, the uncertainty is caused by a difficult to detect color, white and yellow. To resolve this, the car could ask a question regarding the location of the object, in front or on the right.

\subsection{Attribute Prediction}
\label{sect:result_attr_pred}
We measure the accuracy of our proposed attribute predictors in {\color{black}Subsection \ref{sect:attribute_predictor}} for the
three different types of attributes: color, action, and spatial location.
\begin{table}[pth]
    \centering
    \begin{tabular}{|c|c|c|c|}
    \hline
        Model & Attribute Type & Acc. & Inf. Speed (ms)
        \\
        \hline
        ResNet-152  & Action    & 0.731 & 40.74         \\
        ResNet-152  & Color     & 0.719 & 40.74    \\\hline
        NN          & Location  & 0.778 &  0.34   \\
        DT          & Location  & 0.794 & 1.51      \\
        RF          & Location  & 0.783 & 5.17  \\
        SVM         & Location  & 0.754 & 103.19 \\
        RBF SVM     & Location  & 0.779 & 207.65 \\
        LR & Location & 0.757 & 1.48 \\
        \hline
    \end{tabular}
    \caption{Attribute Prediction Results.}{Results of the attribute prediction on the Talk2Car-Expr validation set, with NN a two layer Neural Network, N. Neigh. is Nearest Neighbour, DT is the Decision Tree, RF is Random Forest, SVM is Support Vector Machine, RBF SVM is a SVM with a Radial Basis Function (RBF) Kernel, and LR is Logistic Regression. Inference time is measured on a RTX TITAN for the ResNet-152 models. For the other models, we measure their inference times on a Intel(R) Xeon(R) Silver 4208.}
    \label{tab:predicting_attributes}
\end{table}
From the results shown in Table~\ref{tab:predicting_attributes} we see that predicting locations is easier than predicting colors or actions. Predicting actions is the most challenging task as it involves reasoning about movement and pose from a 2D image.

We see that for predicting the location, the decision tree has the highest accuracy. Although the Neural Network has a faster inference time, we decide to use the decision tree for our next experiments as the difference in inference time is relatively small.
We display the best result obtained with a multitask network that jointly predicts action and color. %
The single task networks had fairly similar results.

For the location neural network, we train using Adam with a learning rate of 0.0003 and a batch size of 16.
The multitask network to jointly predict action and color was trained with Adam with a learning rate of $5 * 10^{-5}$, batch size 16, and weight decay of $0.0001$.

\subsection{Referring Expression Generation - Quantitatively}
For our referring expression generation, we perform two experiments: A quantitative one and a qualitative one.
The former
is performed by generating expressions of objects on the Talk2Car-Expr test set by giving each model an image together with the bounding box of an uncertain object.
The generated expressions are evaluated with the metrics explained in {\color{black}Subsection \ref{sect:language_metrics}}: METEOR, ROUGE-l, and BLEU-4.

\paragraph{Hyperparameter Search for \gls{reg_model}}
We had to finetune multiple settings to achieve optimal results for our CNN-LSTM and \gls{reg_model} models.
We train with three different random seeds during the hyperparameter search for a maximum of 75 epochs and select the best epoch based on METEOR \citep{denkowski2014meteor}.

Based on the validation set, we found that the Adam optimizer with a learning rate of $5 * 10^{-6}$ performs best \citep{kingma2014adam}, the optimal fully connected layer sizes are 512, except for the attention layers where it is 256.
We use a batch size of 16, and during inference, we use a beam search with a beam-size of 10, where the generation is started with a start-of-sentence token.
In the next experiments, we train with a maximum of 125 epochs and select the best epoch using METEOR.

All our expressions are cut to a maximum length of 15 tokens with an additional start-of-sentence token and an end-of-sentence token. All tokens are converted to pre-trained Glove embeddings \citep{pennington2014glove}.
Tokens without a pre-trained embedding that occur more than five times are randomly initialized. The remaining tokens are cast to unknown tokens. We also test with a pre-trained BERT model for extracting the embeddings \citep{devlin2018bert}.
However, this significantly decreased performance and inference speed.

\paragraph{Tuning the Loss}
For the finetuning of the parameters for the \gls{mmi-mm}-loss and the switch-loss on the validation set, we report our findings in Table~\ref{tab:mmi_tune}~and~\ref{tab:switch_tune}, respectively. The results are given for the METEOR metric.

For the \gls{mmi-mm}-loss we had to tune the margin used in the margin $M$ from Eq.~\ref{eq:loss_mmi} and its weight \gls{m:weight_mmi}.
For the \gls{mmi-mm}-loss, we see that a small margin is advantageous and that a higher margin decreases the model's performance.
It is beneficial to use this loss, though only with a small weight.
With the weight \gls{m:weight_mmi} set to $0.1$ we achieve the best results.
Therefore, for all further experiments we set the weight \gls{m:weight_mmi} to $0.1$ and the margin $M$ to $0.1$.

Opposed to the \gls{mmi-mm}-loss, we find that when the switch and its loss are used, it is essential to set its weight high.
We decided not to make it larger than one since the main objective is still the generation of correct referring expressions. For all future experiments with the switch, we set its weight \gls{m:weight_switch} to $1$.
Furthermore, when tested with several loss functions (Cross-Entropy, Smooth-L1, and MSE), but found that the MSE loss, as described in Eq.~\ref{eq:loss_switch}, performs best.

\begin{table}[pt]
\centering
\begin{tabular}{|l|l|llll|}
\hline
\multicolumn{2}{|c|}{\multirow{2}{*}{METEOR ($\uparrow$)}} & \multicolumn{4}{c|}{Margin}       \\ \cline{3-6}
\multicolumn{2}{|c|}{}                        & \multicolumn{1}{c}{0.1} & \multicolumn{1}{c}{0.2} & \multicolumn{1}{c}{0.5} & \multicolumn{1}{c|}{1} \\ \hline
\multirow{4}{*}{weight}         & 0           & \underline{0.2484} & \underline{0.2484} & \underline{0.2484} & \underline{0.2484} \\
                                & 0.1         & \textbf{0.2512} & 0.2482 & 0.2482 & 0.2471 \\
                                & 0.5         & 0.2483 & 0.2469 & 0.2469 & 0.2447 \\
                                & 1           & 0.2464 & 0.2461 & 0.2437 & 0.2424 \\ \hline
\end{tabular}
\caption{MMI Loss Tuning.}{Results for finding optimal weight for the MMI loss, as well as the margin value. Results for validation split with the \gls{reg_model}-Full. The best score is bold with the second best underlined. Since the margin has no effect with a weight of zero, the same score is reported for all.}
\label{tab:mmi_tune}
\end{table}

\begin{table}[pt]
\centering
\begin{tabular}{|l|l|lll|}
\hline
\multicolumn{2}{|c|}{\multirow{2}{*}{METEOR ($\uparrow$)}} & \multicolumn{3}{c|}{switch\_loss}     \\ \cline{3-5}
\multicolumn{2}{|c|}{}                        & ce     & mse             & smooth\_l1 \\ \hline
\multirow{4}{*}{weight}         & 0.1         & 0.1904 & 0.2490          & 0.2492    \\
                                & 0.5         & 0.1901 & 0.2529          & 0.2522    \\
                                & 0.9         & 0.1910 & \underline{0.2538}    & 0.2524    \\
                                & 1           & 0.1911 & \textbf{0.2544} & 0.2529    \\ \hline
\end{tabular}
\caption{Switch Loss Tuning.}{Results for finding optimal loss function and the weight for the Switch Loss. Results for validation split with the \gls{reg_model}-Full. The best score is bold, with the second best underlined. Since the margin has no effect with a weight of zero, the same score is reported for all. }
\label{tab:switch_tune}
\end{table}

\begin{table*}[th]
    \centering
    \begin{tabular}{|c|c|c|c|c|}
\hline
Method & METEOR ($\uparrow$) & ROUGE-l ($\uparrow$) & BLEU-4 ($\uparrow$) & Inf. Speed (ms)\\
\hline
SLR~\citep{yu2016joint} & 0.268 & 0.597 & 0.245 & 663.90\\
SR~\citep{tanaka2018generating} & 0.272 & 0.603 & 0.261 & 1308.06\\
\hline
CNN-LSTM-box & 0.22 & 0.549 & 0.208 & 152.6\\
CNN-LSTM-full & 0.228 & 0.555 & 0.212 & 150.1\\
\gls{reg_model} & 0.247 & 0.582 & 0.231 & 166.8\\
\gls{reg_model}-hot & 0.267 & 0.615 & 0.261 & 162.5\\
\gls{reg_model}-att & 0.286 & 0.642 & 0.287 & 152.5\\
\gls{reg_model}-full & 0.274 & 0.631 & 0.278 & 167.3\\
\hline
\gls{reg_model}-hot+cls & \textit{0.286} & 0.646 & \textit{0.302} & 178.7\\
\gls{reg_model}-att+cls & \textit{0.286} & \textit{0.647} & 0.300 & 145.3\\
\gls{reg_model}-full+cls & 0.285 & 0.644 & 0.295 & 163.6\\
\hline
\gls{reg_model}-hot+switch & 0.26 & 0.608 & 0.259 & 171.1\\
\gls{reg_model}-att+switch & 0.28 & 0.637 & 0.294 & 189.2\\
\gls{reg_model}-full+switch & 0.27 & 0.627 & 0.283 & 175.5\\
\hline
\gls{reg_model}-hot+cls+diff & \underline{0.288} & \textbf{0.652} & \textbf{0.306} & 217.9\\
\gls{reg_model}-att+cls+diff & \textbf{0.29} & \underline{0.651} & \underline{0.301} & 197.0\\
\hline
    \end{tabular}
    \caption{Quantitative Evaluation for Referring Expression Generation.}{Quantitative evaluation of our proposed models and state-of-the-art baselines on the Talk2Car-Expr \textbf{test} set. CNN-LSTM-box only uses the ResNet-152 box features, and CNN-LSTM-full additionally integrates ResNet-152 features for the entire image. The \glspl{reg_model} makes use of the object label (\gls{reg_model}), with optionally the attributes and object label as a one-hot vector (\gls{reg_model}-hot), optionally an attention over the word embeddings for the attribute and the class label (\gls{reg_model}-att), or both of them (\gls{reg_model}-full). All options can be extended with the class label embedding inputted in the LSTM of the decoder at every timestep (+cls) or with a trained switch that decides if attributes should be forced in the expression (+switch). AttrExpr-full is the model using the full set of features: attention with both attributes and class embedding, the image feature vector, box feature vector, and the hot-vectors for attributes and classes. Inference speed is measured on an RTX TITAN. To indicate the best, second, and third best scores we use \textbf{boldface} and \underline{underline} and \textit{cursive}, respectively.}
    \label{tab:quantitative_evaluation_description}
\end{table*}

\subsubsection{Results of the Referring Expression Generation}
Table~\ref{tab:quantitative_evaluation_description} shows the results for different combinations of settings for generating expressions on the test set of Talk2Car. The results are grouped into
the state-of-the-art baselines, our model variations of CNN-LSTM and \gls{reg_model}, our model variations plus the extra class label embedding, our model variations plus the switch for forcing attributes, our model variations with both the class label embeddings and the switch, and finally our model variations with the difference features.

For the baselines, we find that the SR model by \citet{tanaka2018generating} performs best across all metrics. However, the inference time is more than twice as long compared to the SLR model by \citet{yu2016joint}, which only performs a few points lower on some of the metrics (e.g., ROUGE-l).
Both baselines also have a way to re-rank multiple simultaneous generated expressions by passing them to either the \textit{Listener} in the case of SLR \citep{yu2016joint} or the \textit{Reinforcer} in the case of SR \citep{tanaka2018generating}.
The expressions are then ranked according to the generated scores by \textit{Listener} or \textit{Reinforcer}.
We found that this did not improve the results on Talk2Car.

For the implementations of our models, we note that \gls{reg_model} performs much better than the simpler CNN-LSTM. This observation confirms our hypothesis that guidance by the attributes of the object aids in describing it. More specifically, when comparing \gls{reg_model} with its variations (-hot, -att, and -full), we note that the attributes and the class label give a great improvement over just using the global image feature \gls{m:f-img}, the bounding box feature \gls{m:f-box}, and the distance count \gls{m:f-cnt}.

Interestingly, according to most metrics the best variations of the \gls{reg_model} model are these that use the additional class label embedding (+cls) on top of \gls{reg_model}-hot or \gls{reg_model}-att. This indicates that using the full model adds unneeded complexity and hurts performance across many of the metrics.
We note that between the models that do not use the class label embedding as input, the \gls{reg_model}-hot is one of the worst-performing variations. This indicates how important the class label is for generating the referring expressions.
Training the switch and forcing attributes are not as beneficial as expected. We expect this has to do with the same complexity issues mentioned before when using the full model and the model becoming too complex to train properly.
Finally, Table~\ref{tab:quantitative_evaluation_description} shows an improvement in terms of METEOR, ROUGE-l and BLUE-4
when the difference features (+diff) are added as input (see paragraph \textit{Adding Difference Features (+diff)} in {\color{black}Subsection~\ref{sect:cnn_lstm}}).
These features are obtained by the subtraction of the object feature from the features from the other objects.
This shows that this kind of delta information about the other objects is beneficial for generating good referring expressions.

\begin{figure*}[ht]
\centering
\begin{subfigure}{0.49\linewidth}
\centering
\includegraphics[width=\linewidth]{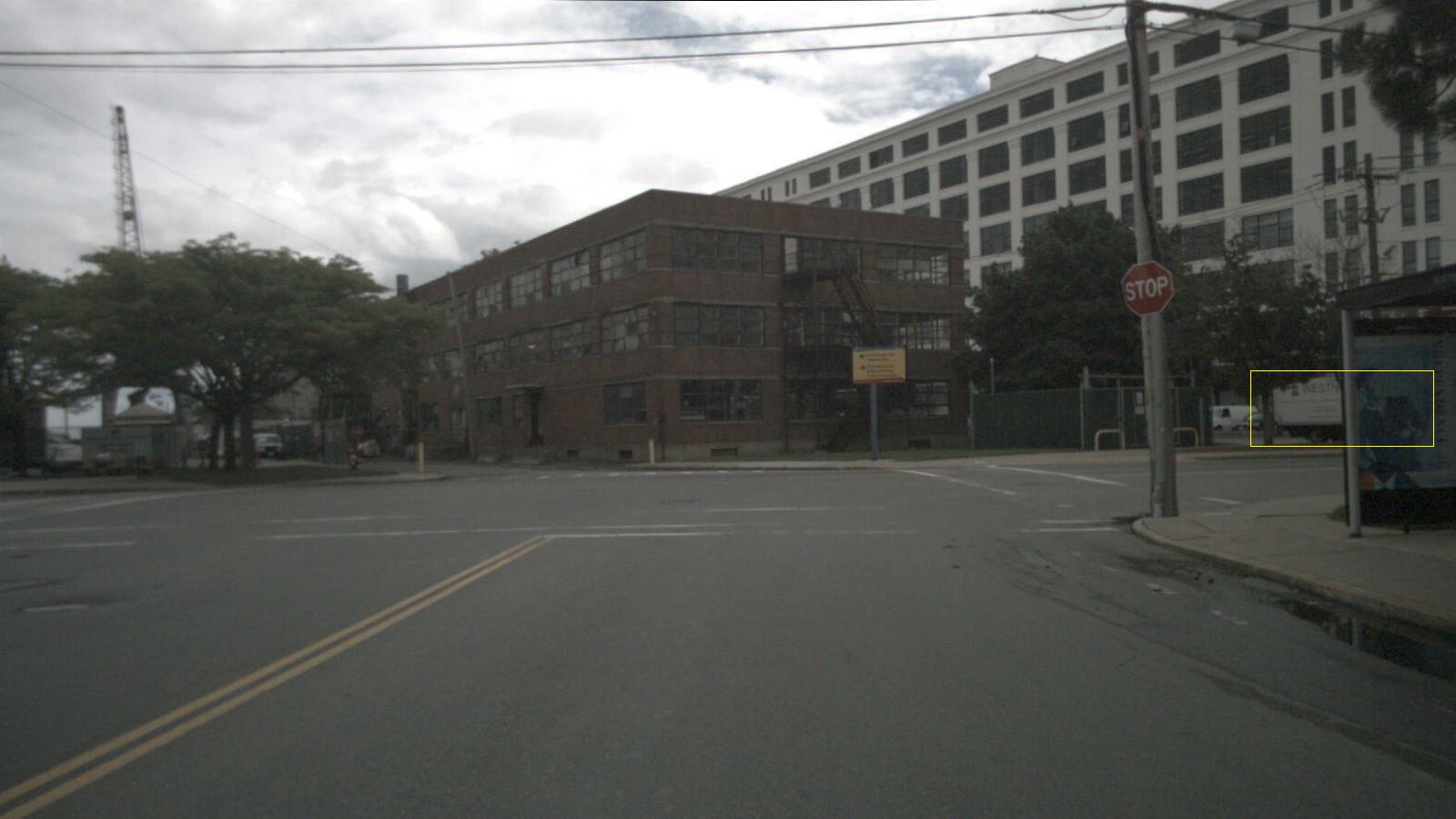}
\caption{Positive example for \gls{reg_model} compared to baselines.}{
\begin{tabular}{rl}
    \textbf{SLR} &  {\color{red}first pushable container on right}\\
    \textbf{SR} & {\color{red}first construction worker on right}\\
    \textbf{A-REG-att+cls+diff} & {\color{darkgreen}first truck on right} \\
    \textbf{A-REG-hot+cls+diff} & {\color{darkgreen}first truck on right}
\end{tabular}
}
\label{fig:gen_expr_succes7}
\end{subfigure}
\hfill
\begin{subfigure}{0.49\linewidth}
\centering
\includegraphics[width=\linewidth]{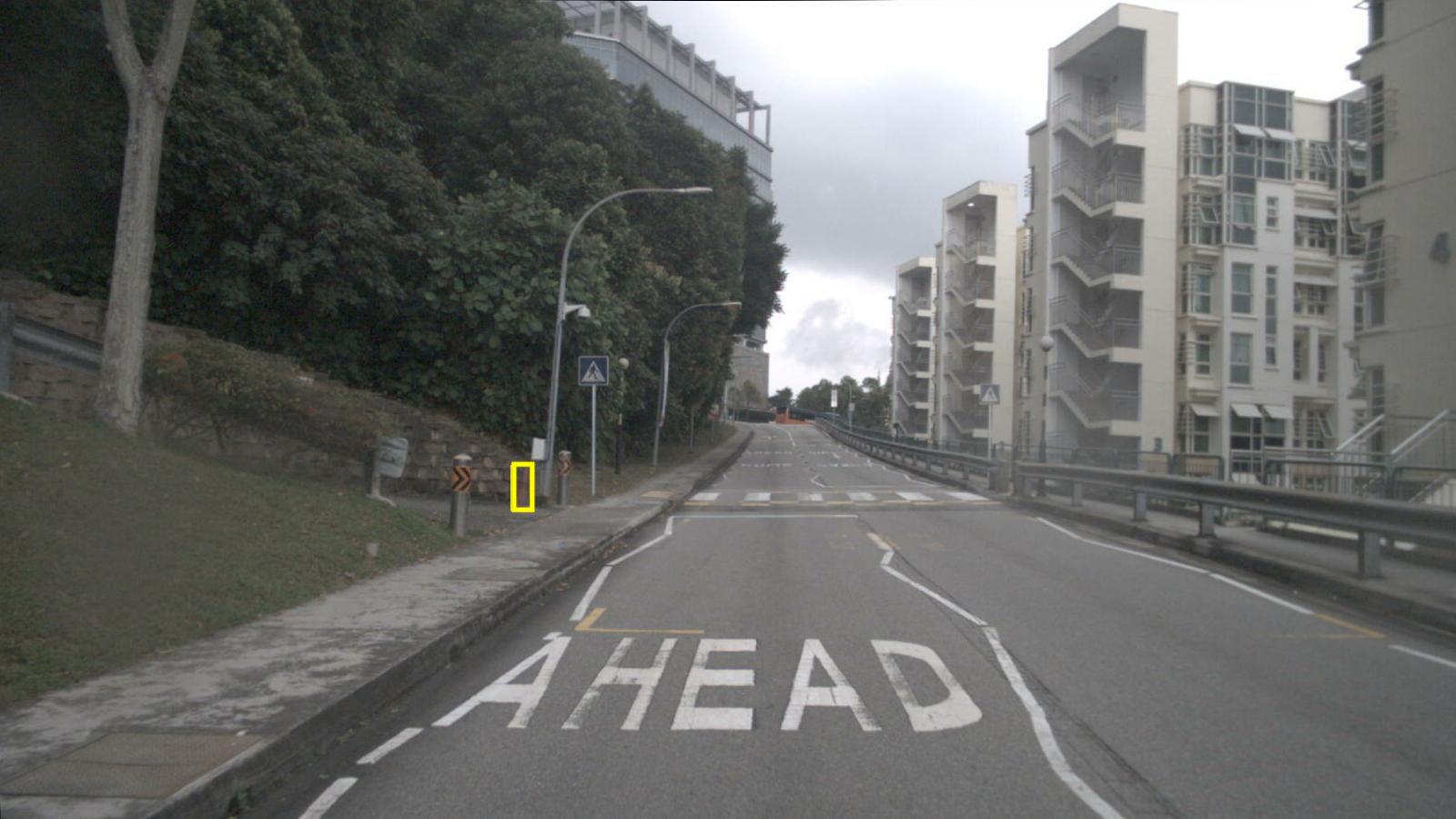}
\caption{Positive example for \gls{reg_model} compared to baselines.}{
\begin{tabular}{rl}
    \textbf{SLR} &  {\color{red}first adult on left}\\
    \textbf{SR} &  {\color{red}first adult on left}\\
    \textbf{A-REG-att+cls+diff} &   {\color{darkgreen}the first orange traffic cone on the left}\\
    \textbf{A-REG-hot+cls+diff} &  {\color{darkgreen}the first orange traffic cone on the left}
\end{tabular}
}
\label{fig:gen_expr_succes3}
\end{subfigure}
\caption{Examples of generated expressions}{Examples of generated expressions for the state-of-the-art Baselines SLR and SR, and the two top performing \acrshort{reg_model} variations, \acrshort{reg_model}-att+cls+diff and \acrshort{reg_model}-hot+cls+diff. {\color{black}We indicate in {\color{red}red} the wrong expressions and in {\color{darkgreen}green} the correct ones.}
Best viewed in color.}
\label{fig:paper_examples_expr}
\end{figure*}

Overall the best implementations of the \gls{reg_model} model outperform the state-of-the-art models across all metrics.
This improvement is most likely due to the use of specifically trained attribute predictors for the Talk2Car task and directly fine-tuning the model parameters for the same task. Furthermore, the inference time is greatly reduced due to a simpler model design.
Two example images with a referred object are shown, with in its caption the expressions generated by two of our top performing models (\gls{reg_model}-hot+cls+diff and \gls{reg_model}-att+cls+diff) and the two state-of-the-art models (SLR and SR), see Figure~\ref{fig:gen_expr_succes7} and Figure~\ref{fig:gen_expr_succes3}. For more examples we refer the reader to \ref{app:examples_generated_expressions}.

Although the automated evaluation with metrics such as METEOR, ROUGE and BLEU, gives a great indication of the quality of the metrics, it is hard to determine whether they measure how well the expressions actually disambiguate the objects in the image.
For instance, we note that our \gls{reg_model} tend to predict somewhat similar referring expressions. We believe that this can be resolved with better attribute predictors and a larger variety of attributes.

\paragraph{Human Evaluation}

\begin{table*}[t!h]
\centering
\begin{tabular}{|c|c|c|c|c|c|}
\hline
\specialcell{Uncertainty\\model} & REG model & \specialcell{Human\\Agreement ($\uparrow$)} & Acc ($\uparrow$) & F1 ($\uparrow$) & %
SingleExpr ($\downarrow$) \\ \hline
\multirow{4}{*}{Ens\_4 + EV}      & SR                 & 0.40 & 0.68 & 0.70 & 0.10\\
                                  & SLR                & 0.41 & 0.69 & 0.70 & 0.09\\
                                  & A-REG-hot+cls+diff & 0.53 & 0.64 & 0.70 & 0.25\\
                                  & A-REG-att+cls+diff & 0.56 & 0.70 & 0.75 & 0.22\\ \hline
\multirow{4}{*}{Ens\_5 + CF + EV} & SR                 & 0.45 & 0.69 & 0.70 & 0.12\\
                                  & SLR                & 0.52 & 0.65 & 0.72  & 0.10\\
                                  & A-REG-hot+cls+diff & 0.59 & 0.68 & 0.69 & 0.42\\
                                  & A-REG-att+cls+diff & 0.57 & 0.77 & 0.77 & 0.37\\ \hline
\end{tabular}
\caption{Human Evaluation Results.}{Human Evaluation using several metrics. For the human agreement we use the Krippendorf Alpha. Acc is the accuracy of correct assigned expressions to objects. SingleExpr is the fraction of images where the used REG model generates the same expression for all uncertain objects.}
\label{tab:human_eval_expr}
\end{table*}

For our REG models, we also performed a human evaluation using Amazon Turk.
We gave the workers an image with a referred object indicated with a bounding box and a set of generated expressions by the used referring expression models in this paper.
These expressions are presented as described in {\color{black}Subsection~\ref{sect:generating_questions}}, where they have to indicate for each object expression \textbf{[expr $o^{n}$]} if it corresponds with the indicated object.
The set of generated expressions was created by using $Ens_4+EV$ or $Ens_5+CF+EV$ as uncertainty detection models, where images were kept if there is exactly one bounding box that indicates the referred object amongst the uncertain objects, to avoid too many overlapping boxes.
This results in 534 uncertain situations for $Ens_4+EV$ and 628 for $Ens_5+CF+EV$.

The results are shown in Table~\ref{tab:human_eval_expr}.
For most metrics, we see that the results are better for \gls{reg_model} compared to the state-of-the-art models.
We note that this is not the case for the \textit{SingleExpr}, indicating that multiple objects received the same expression. However, when looking at the examples in Figure~\ref{fig:paper_examples_expr} the difference between the state-of-the-art baselines is that \gls{reg_model} is %
able to correctly identify the objects in the box. Therefore, we believe that this issue is more acceptable for the end-user.

\subsection{Final Result}
\begin{figure*}[th!]
    \centering
    \includegraphics[width=\linewidth]{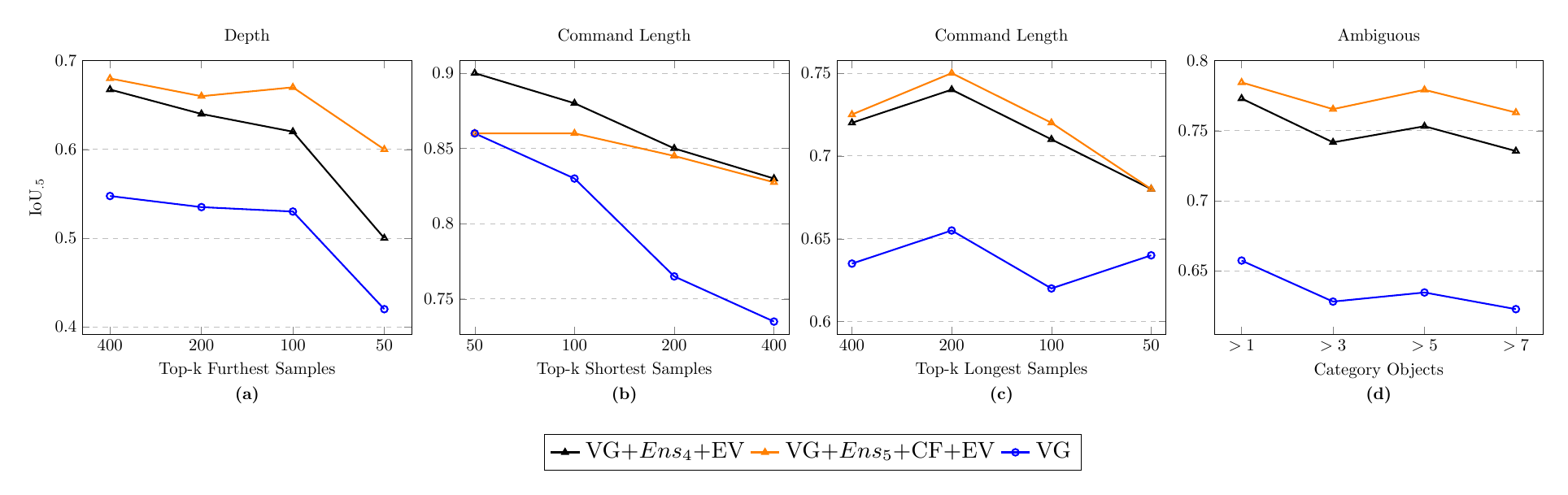}
    \caption{Results on Talk2Car subsets.}{
    In this figure, we plot the $IoU_{.5}$ score of the used VG model ({\color{black}Subsection \ref{sect:CU_model}}) and the VG model in conjunction with the top two uncertainty methods from {\color{black}Subsection \ref{sect:exp_jointly_detecting_uncertainty}}: $Ens_4$+EV and $Ens_5$+CF+EV. The former model means that we have an ensemble of four VG models using ensemble voting (EV) to determine the uncertain objects. The latter indicates that we have five VG models that first use class filtering (CF) and then EV for detecting the uncertain objects. At the top of each plot, we give the name of the subset. Each plot also shows the easy examples on the left and increases the difficulty when moving along the plot's x-axis. Best viewed in color.
    }
    \label{fig:final_accuracy_plot}
\end{figure*}
To finish this paper, we investigate the accuracy of the full \gls{pipeline}.
We hypothesize that the final accuracy of \gls{pipeline} is equal to the theoretical accuracy column ($Th.IoU_{.5}$) from Table \ref{tab:results_uncertainty_experiment_validation_set} regardless of the generated question for the uncertain objects.
We argue that the passenger who has given the command will always select the object they refer to if it is part of the selected uncertain objects.
Following this reasoning, we can say that the overall improvement of using the \gls{vg} model with \gls{pipeline} compared to only using the \gls{vg} model is a 9\% absolute increase whenever we use $Ens_4 + EV$ or $ Ens_5 + CF + EV$ to detect uncertainty.

We are now interested to see how \gls{pipeline} fares compared to the best performing \gls{vg} model from Table \ref{tab:influence_of_number_of_objs} without the uncertainty system on the four challenging subsets present in Talk2Car ({\color{black}Subsection \ref{sect:dataset}}).
As discussed above, Figure \ref{fig:final_accuracy_plot} shows that the combination of the \gls{vg} model and the proposed uncertainty detection models largely improves the $Th.IoU_{.5}$ metric when confronted with difficult visual situations (for the depth subset we witness an increase of 12.1\%), with long language commands (for the longest sentences subset we witness an increase of 7.8\%) and with referred objects in the commands that have an ambiguous interpretation in the visual scene (for the ambiguous subset we witness an increase of 12.6\%), while for easier situations the increase is less pronounced (for the short sentences subset we witness an increase of 5.9\%).
These results show that for the difficult situations, our uncertainty detection models are very beneficial.
Finally, the overall inference speed of \gls{pipeline} is around 286.99ms in case of using the $Ens_4 + EV$ uncertainty detection method or 336.05ms in case of using $Ens_5 + CF + EV$ as the uncertainty detection method and A-REG-hot+cls+diff as REG model.
Using the state-of-the-art baseline SLR as the REG model, this would be 732.99ms and 782.05ms. For SR, the total inference speed would be or 1447.55ms and 1495.66ms.

{\color{black} %
\subsection{Expected Requirements and Maintenance for Implementing the \acrfull{pipeline} Pipeline}
We assume the use of a modern self-driving car, which is already equipped with cameras and (possibly) LIDAR or radar sensors. Furthermore, most modern cars have microphones and a speech recognition function, which can be used to give commands regarding music and text messaging while driving.

We first discuss the capital expenditure of our system.
We believe that adding our pipeline to a self-driving car would not require much work. We could directly couple our system with existing infotainment systems that can be controlled through speech (i.e., BMW ConnectedDrive\footnote{\url{https://www.bmw-connecteddrive.be/app/index.html#/portal}}, Ford Sync\footnote{\url{https://www.ford.com/technology/sync/}}, ...).
To do so, a system must be added that detects when a command is given regarding controlling the car's internal systems, or a command regarding the surrounding of the car. This could be achieved with a simple classifier choosing between the two options. If the latter is detected, the command is forwarded to our pipeline. A pre-trained model is loaded into the cars internal computer, such that the model is small and quick to run. As shown by the results, the complete inference time from the entire \gls{pipeline} pipeline with our proposed REG model is very low, always staying below 400 ms. The internal computer must have enough capacity to maintain a low inference time.

For the operating expenditure we need
to make sure that the model is capable of improving and solving future
issues.
It is important to roll out updates of the pre-trained CU and A-REG models.
Compared to
current interactions of a driver with a Tesla car (e.g., during braking or steering), the interactions of a passenger of the self-driving car can be logged
and used as feedback to further retrain and fine-tune our system.
These updates can then be published
in a similar fashion as Tesla's regular over-air updates\footnote{\url{https://www.tesla.com/support/software-updates}}.
When technology advances and new discoveries are made, a major update can be released to the car, introducing a novel CU or REG model.
}

\section{Conclusion}
\label{sect:conclusion}
In this paper, we have proposed the \acrfull{pipeline} for a self-driving car. \acrfull{pipeline} augments a Visual Grounding (VG) model with the ability to (1) detect if a given natural language command leads to uncertain situations and (2) finding the objects causing the uncertainty.
The best method for these two abilities is to use an ensemble of VG models in conjunction with Ensemble Voting (EV) and optionally Class Filtering (CF).
Our contribution for detecting uncertainty lies in (1) evaluating many different methods and their combination for detecting said uncertainty and (2) proposing a novel set of constraints tailored for a self-driving setting.
\gls{pipeline} has a visual output or, in addition to the former, it can also present the passenger a generated question that describes the uncertain objects.
For this purpose, we have developed a new referring expression model that we called \gls{reg_model}.
It is designed as a robust two-layer LSTM referring expression generator that can efficiently use an object's attributes for the textual output.
These attributes, specifically tailored for the Talk2Car dataset, include the bounding box, distance count, color, location, action, and the class label.
{\color{black} If one would like to change the set of attributes, this can simply be done by retraining %
the attribute predictor based on data annotated with the new set of attributes and by retraining the A-REG model. %
}
Our method stands in contrast to the single layer network of the state-of-the-art baselines.
The first layer processes all the global information (such as the image representation, object representation, and one-hot encodings for the attributes). The second layer receives all the inputs required to predict the next expression token (such as the object properties, embeddings for the class label, and the attributes).
In our experiments, we show that by using \gls{pipeline}, a \gls{m:overall-increase-VG} absolute increase in terms of $IoU_{.5}$ is achieved compared to the VG model without \gls{pipeline} on the Talk2Car dataset.
Additionally, we also show that \gls{pipeline} helps the most in challenging situations such as objects being far away or multiple objects of the same class.
In these cases, our \gls{pipeline} increases the $IoU_{.5}$ of the used VG system $IoU_{.5}$ by 12.1\% and 12.6\% respectively.
We show that \gls{reg_model} for referring expression generation beats existing state-of-the-art models on the Talk2Car dataset by leveraging the previously mentioned attributes.
With all these features combined the state of the art is relatively outperformed with up to \gls{m:meteor-relative} METEOR and \gls{m:rouge-relative} ROUGE-l but at a nearly three times faster inference speed.

\paragraph{Future Work}
We identify several exciting possibilities as future work.
First, in this work the questions posed to a passenger of the self-driving car are in textual format assuming their translation into speech by a text-to-speech engine.
It would be interesting to study how speech signals could be generated, describing the visual scene's uncertain objects without any intermediate textual step. This might also benefit the latency when generating a question in speech format.
Second, as mentioned in our experiments, we experimented with combining multiple models from an ensemble into one. However, the achieved results were worse than using the models of the ensemble separately.
Concerning power consumption on a self-driving car, using an ensemble of four or five models might be expensive.
Hence, it might be interesting to study how the computations of uncertainty detection and quantification can be put into one model.
{\color{black} Although we demonstrated the validity %
of uncertainty detection %
when understanding natural language commands
in a
visual context,
it might be interesting to investigate whether the proposed method performs well in other tasks executed by
a self-driving car. For instance, in bad weather conditions it is important to detect how uncertain the visual recognition is, so that the car can take appropriate actions (e.g., pull over to avoid an accident). }
Finally, we note that the referring expression models sometimes generate the same sentence for multiple objects in the scene. One reason is the overlap between predicted boxes.
{\color{black} %
Another reason is the level of detail of the object's class and its attribute labels. We believe there is
a trade-off between how detailed the attributes should be, causing longer questions, versus creating short descriptions that can be quickly understood by the passenger.
Future work could
investigate the level of detail of attributes
without them
endangering the safety of the passenger.}

\section*{Acknowledgement}
This project was supported by the MACCHINA project from the KU Leuven with grant number C14/18/065 and has received funding from the European Research Council (ERC) under the European Union's Horizon 2020 research and innovation programme (CALCULUS project Grant agreement No. 788506).
We also wish to thank NVIDIA for providing us with two RTX Titan's XP's for our experiments.

\bibliographystyle{elsarticle-harv}
\bibliography{main}
\appendix
\onecolumn
\section{Conducted Survey}
\label{app:survey}
For creating this application paper, we held a survey to know how people would like to have a self-driving car report back about its uncertainty.
We present the survey results here, and we hope that some of these results could also lead to future work.

\begin{figure}[H]
\centering
\begin{subfigure}[t]{0.303\textwidth}
    \centering
    \includegraphics[scale=.31]{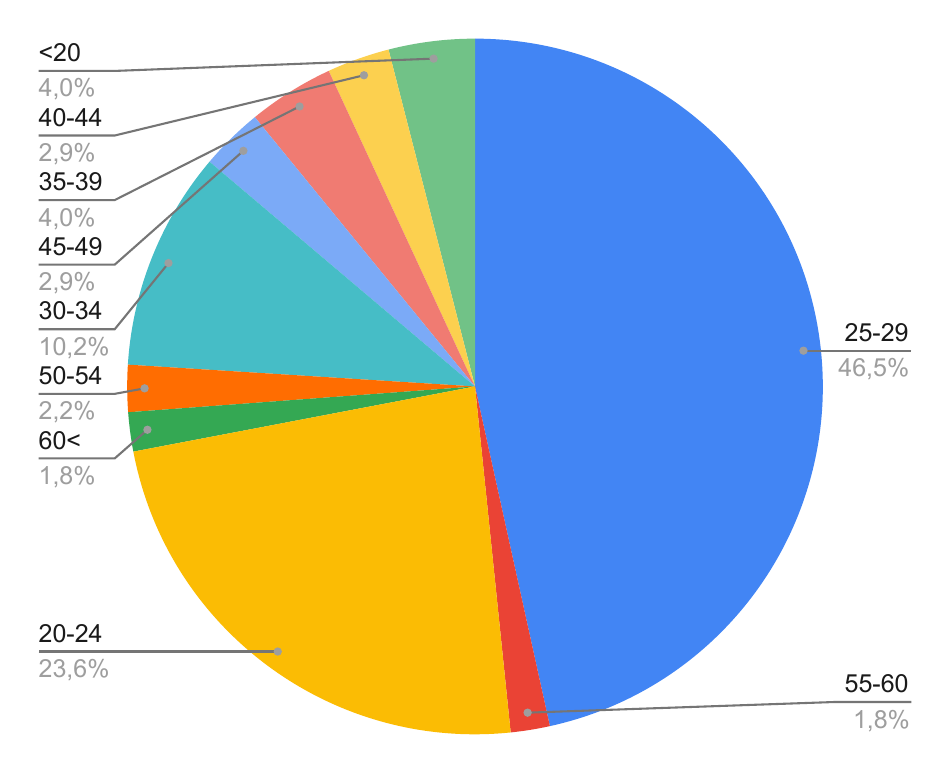}
    \caption{Age of participants from the survey}
    \label{fig:survey_age}
\end{subfigure}
\begin{subfigure}[t]{0.344\textwidth}
    \centering
    \includegraphics[scale=.31]{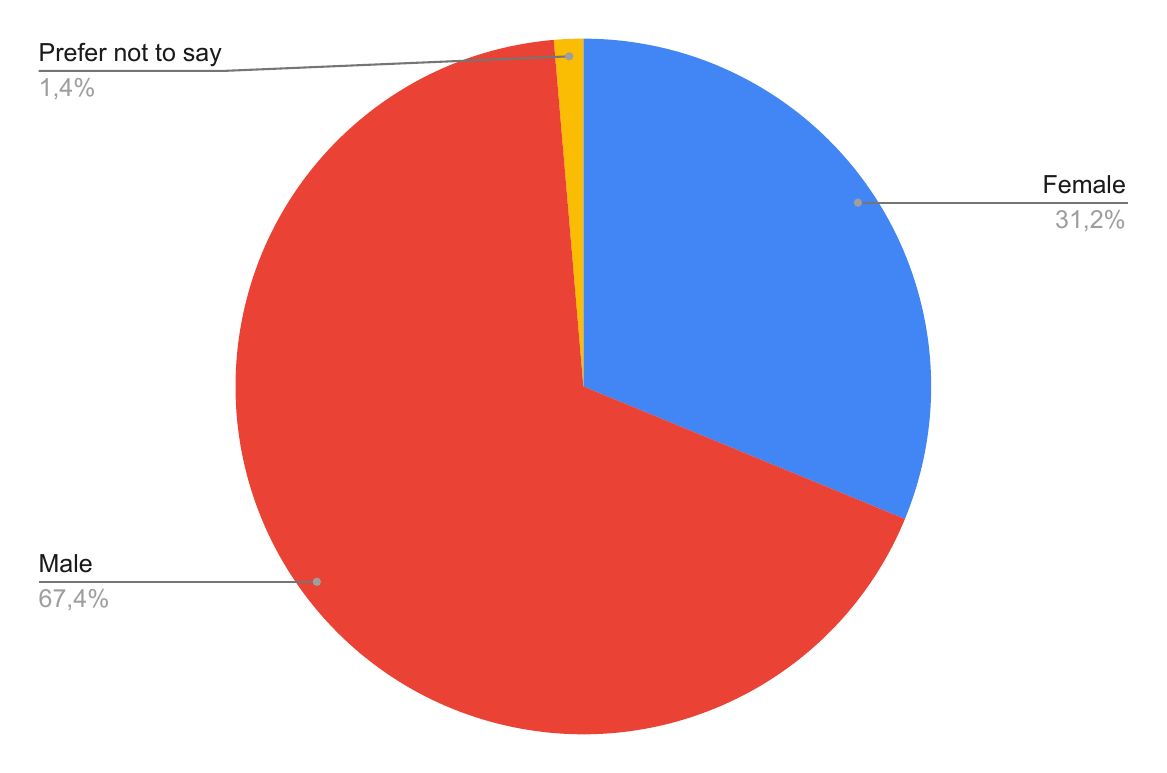}
    \caption{Gender of participants from the survey}
    \label{fig:survey_gender}
\end{subfigure}
\begin{subfigure}[t]{0.344\textwidth}
    \centering
    \includegraphics[scale=.31]{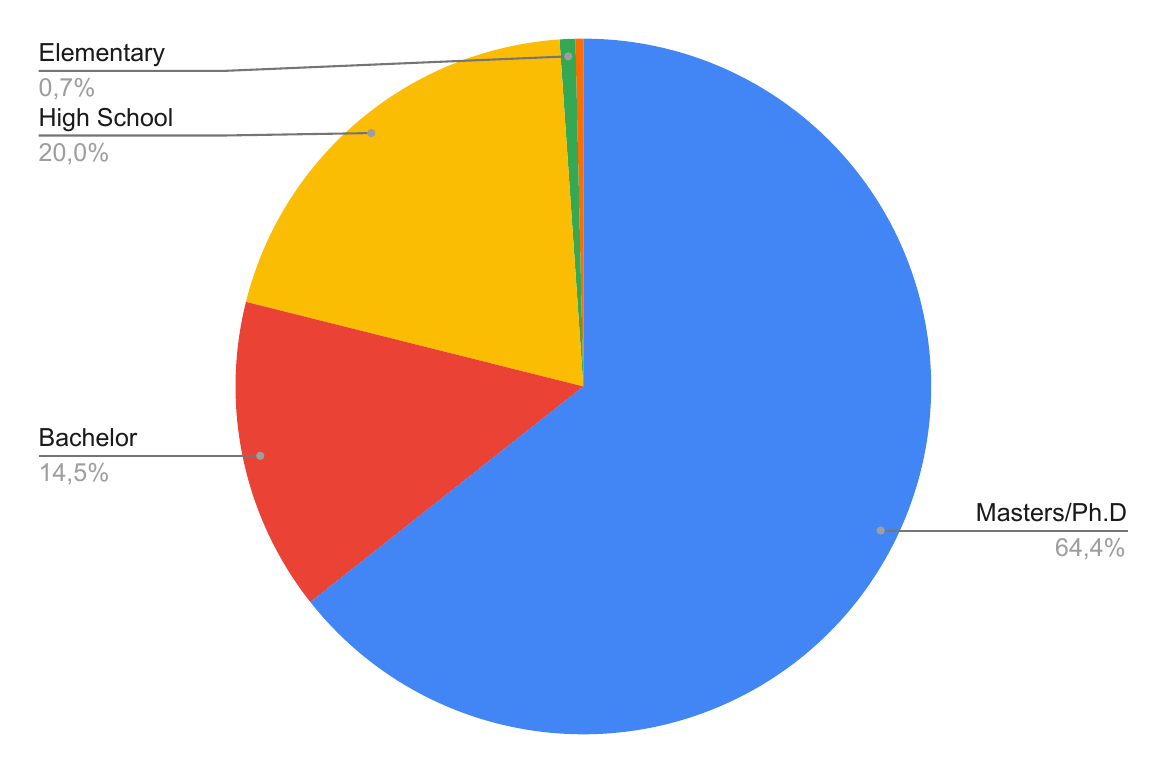}
    \caption{Highest degree of participants from the survey}
    \label{fig:survey_diploma}
\end{subfigure}
\caption{Survey Participants.}{In these plots we show the distribution of participants among different age categories (subfigure a), different genders (subfigure b), and different levels of education (subfigure c). Best viewed in color.}
\label{fig:survey_participants}
\end{figure}

The conducted survey on Social Media (LinkedIn, Facebook, mailing lists, ...) received 254 responses amongst people of different ages, genders, and educations (See Figure~\ref{fig:survey_participants}).

\begin{figure}[H]
\centering
\begin{subfigure}[t]{0.33\textwidth}
    \centering
    \includegraphics[scale=.31]{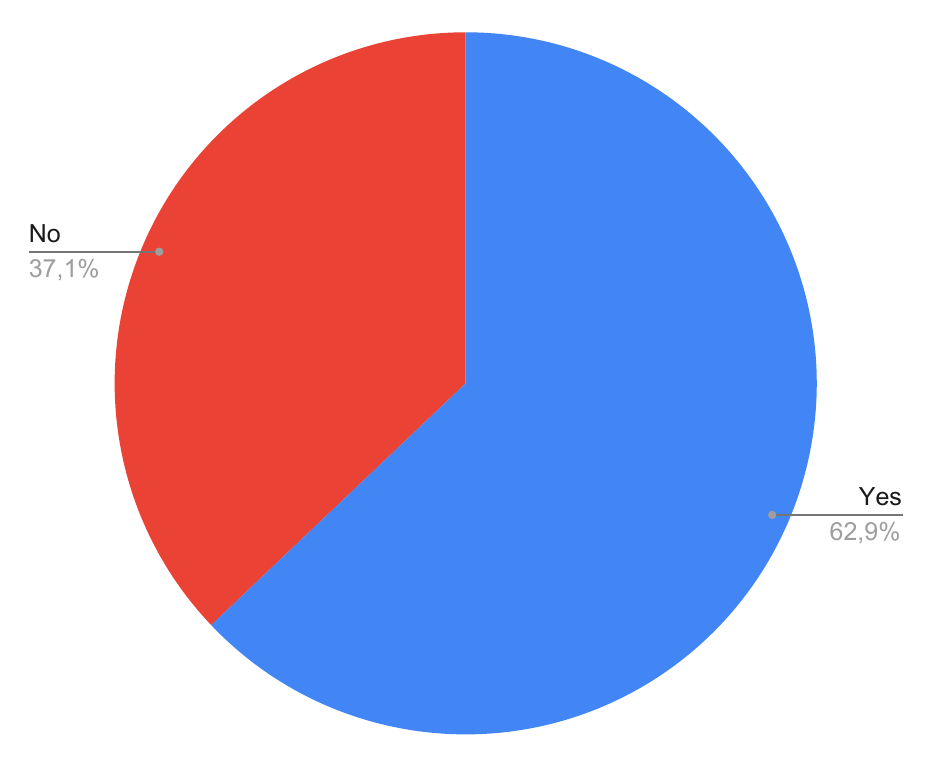}
    \caption{\textbf{Question:} ``Would you trust driving around in a self-driving car?''}
    \label{fig:survey_trust}
\end{subfigure}
\begin{subfigure}[t]{0.33\textwidth}
    \centering
    \includegraphics[scale=.31]{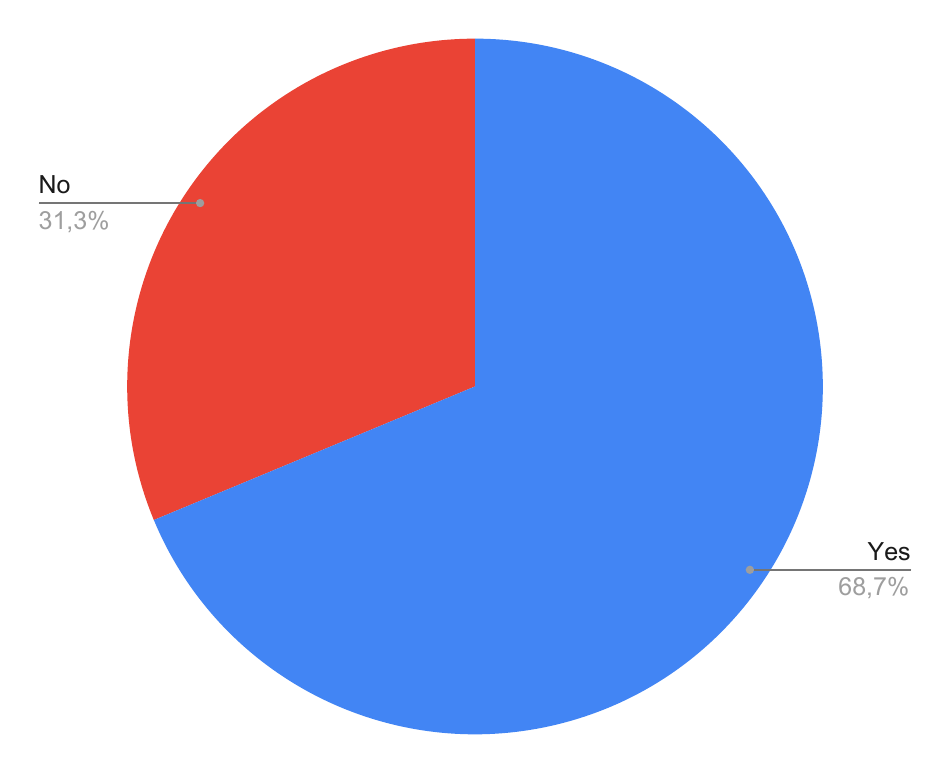}
    \caption{\textbf{Question:} ``If you would be able to give commands to the self-driving car, would you feel more confident driving around in such a car?''}
    \label{fig:survey_trust_command}
\end{subfigure}
\begin{subfigure}[t]{0.33\textwidth}
    \centering
    \includegraphics[scale=.31]{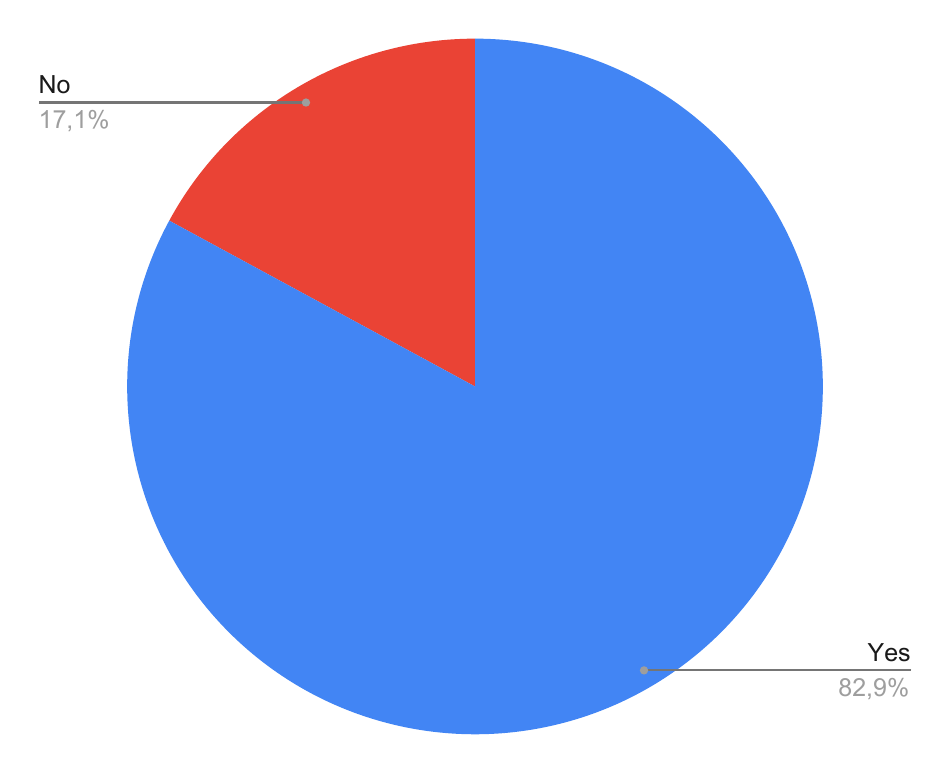}
    \caption{\textbf{Question:} ``Would you use a system that allows you to give commands to a self-driving car? Example commands could be: park in the shade, ...''}
    \label{fig:survey_use_command}
\end{subfigure}
\caption{Results for survey questions regarding trust.}{Results for three questions regarding the trust in the self-driving car with or without option to give commands. Questions are reported in the captions below the subfigures. Best viewed in color.}
\label{fig:survey_trust_compare}
\end{figure}

As shown in Figure~\ref{fig:survey_trust}, we found that a majority does not trust self-driving cars yet. Among these people, we asked for their argument why this is. Below in Table~\ref{tab:survey_responses} is an excerpt from the 91 responses.

We noted that most of the responses have to do with safety concerns and fear of giving away control to an autonomous vehicle. We also note that this often has to do with a lack of knowledge. Furthermore, we found that some of them have concerns regarding the legality and responsibility in case of an accident. Some of these concerns are caused by media and examples of incidents with self-driving cars.

Following these questions regarding trust, we asked whether the ability to give commands to the car, like those in the Talk2Car dataset, would improve this trust. We found that with the ability to give commands the percentage of people who would feel confident, increases to 69.6\% compared to the original 63.2\% ( Figure~\ref{fig:survey_trust_compare}). However, when asking if people would use the ability to give commands, we found that almost all participants (82.6\%) will make use of this, as shown in Figure~\ref{fig:survey_use_command}.

\begin{figure}[H]
\centering
\begin{subfigure}[t]{0.31\textwidth}
    \centering
    \includegraphics[scale=.31]{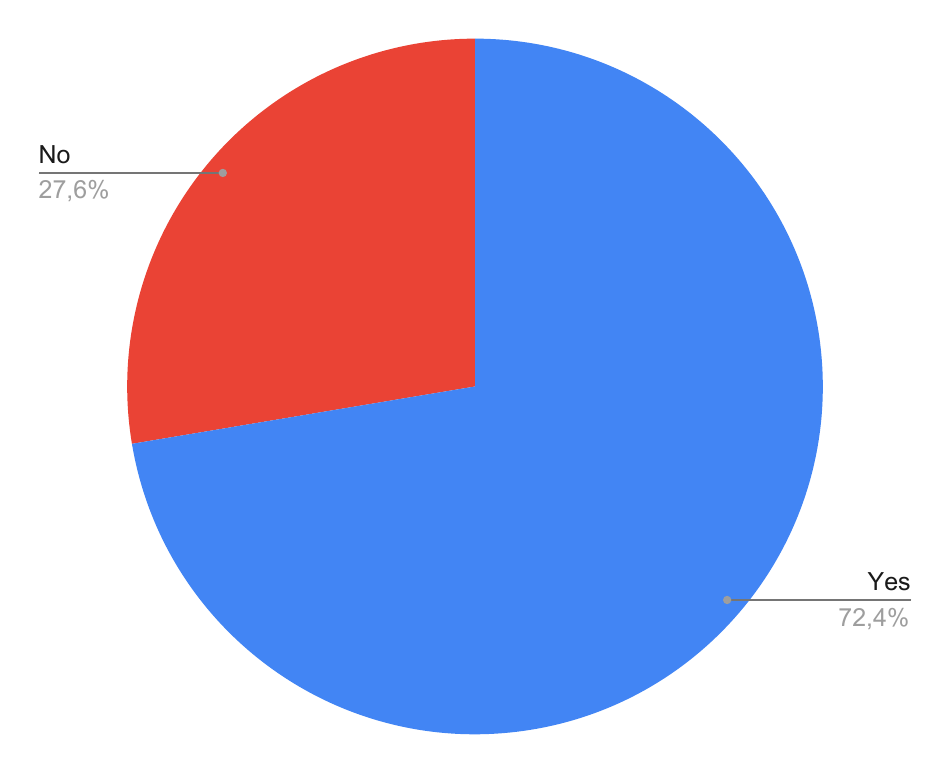}
    \caption{\textbf{Question:} ``Do you want a self-driving car that can respond to you by speech? ''}
    \label{fig:survey_response}
\end{subfigure}
\begin{subfigure}[t]{0.33\textwidth}
    \centering
    \includegraphics[scale=.31]{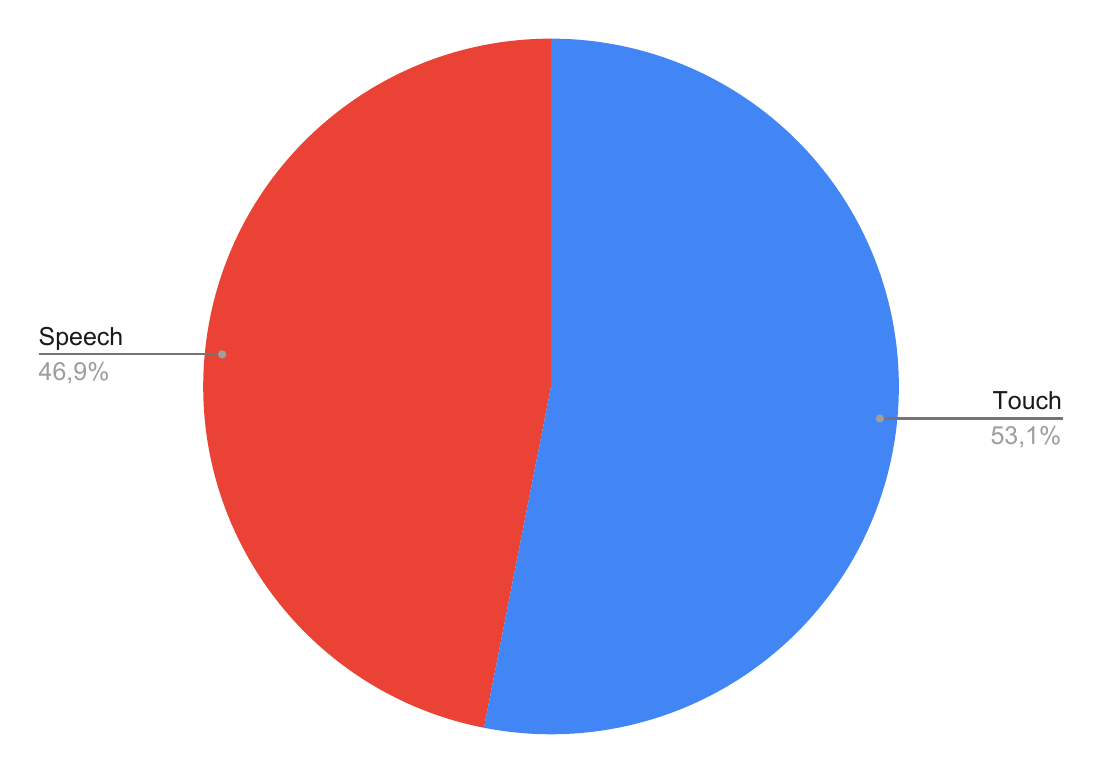}
    \caption{\textbf{Question:} ``Would you prefer to use a touch screen or speech to answer back to the car?''}
    \label{fig:survey_answer_back}
\end{subfigure}
\begin{subfigure}[t]{0.35\textwidth}
    \centering
    \includegraphics[scale=.31]{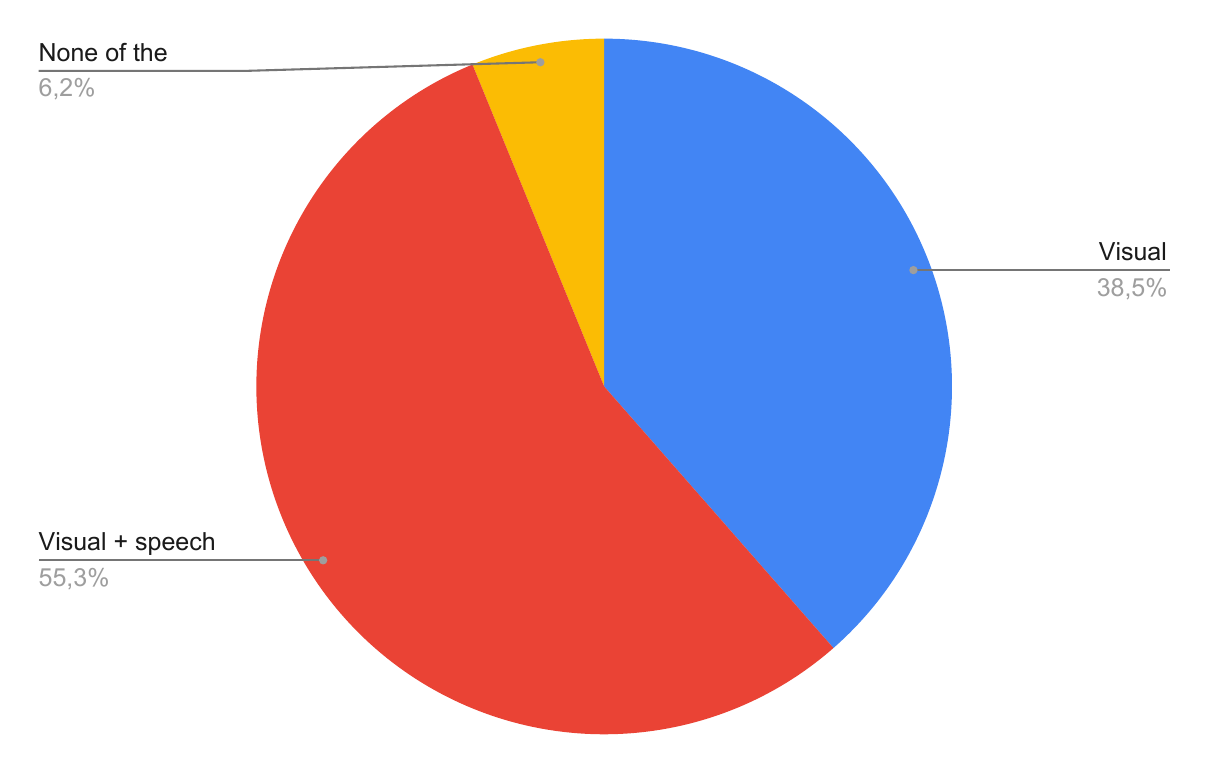}
    \caption{\textbf{Question:} ``You are in a self-driving car. At a particular moment, you give the command "pick up that person." However, there are two persons in front of the vehicle. The car detects this and has the following two options to make you aware of this: (Option 1) It displays the uncertain objects on a touch screen with a rectangular box for each object, or (Option 2) in addition to showing the objects on a touch screen, the car also describes the objects in a question through speech. With which option would you feel more confident in the abilities of the car? If none of the options make you feel more confident, please indicate 'none of the above'.'' Best viewed in color.}
    \label{fig:survey_options}
\end{subfigure}
\caption{Results for survey questions regarding system format.}{Results for three questions regarding the format of the uncertainty resolving system for commands given to the self-driving car. Questions are reported in the captions below the subfigures. Best viewed in color.}
\label{fig:survey_format}
\end{figure}

Finally, we asked a couple of questions regarding the format for uncertainty resolution, with the results shown in Figure~\ref{fig:survey_format}. We found that 71.9\% of the participants would like the car to respond by speech (Figure~\ref{fig:survey_response}). However, when asked how they would prefer the vehicle to report back about its uncertainty, either by only visually showing the uncertain objects or also with a textual/spoken question, we found that only 38.7\% (Figure~\ref{fig:survey_options}) would prefer the former and 54.9\% would prefer the latter. The remaining participants indicated they do not have a preference. Finally, from the survey, we also found that people have a slight preference for responding to the car via a touch screen compared to via a touch screen augmented with speech (Figure~\ref{fig:survey_answer_back}).

\begin{table}[H]
\centering
\begin{tabular}{r|l}
\textbf{Safety} &
\specialcell{It's about trust.  Everything tech is never safe for me.\\
             it will take us years to test self-driving cars so they are at least equally safe as an experienced driver.\\
             Afraid of giving up control and not being certain that the car will be able to respond \\quickly to any given situation. \\ I think the roads have too many exceptions for it to work well. }\\
&\\
\textbf{Control} &
\specialcell{Because I would hesitate to give away all control. I don't trust todays \\technology enough to depend my life on it.\\
             Unless I have 100\% control, I wouldn't feel safe. \\ I am a bit a control freak. \\ I would trust a self-driving car IF I would still be able \\to take over control whenever I want to. }\\
&\\
\textbf{Examples} & \specialcell{People have died in self-driving Teslas.\\
Because of the uber incident.\\
see Tesla crash vids...\\
Reading the current news related to the technology, it seems that we are not quite \\there yet to have a fail-proof system for self-driving cars. Once we are there, \\then I would feel confident.}\\
&\\
\textbf{Human Skills} &
\specialcell{Complexity of human behavior in traffic + too abstract as a concept in this stage.\\
             Computers have not yet achieved the cognitive capabilities of humans.}\\
&\\
\textbf{Responsibility} &
\specialcell{Because if anything happens, I, as the owner of the car would have to take full responsibility.}
\end{tabular}
\caption{Reasons for not trusting self-driving cars.}{An excerpt of some of the answers that participants of the survey gave for not trusting self-driving cars.}
\label{tab:survey_responses}
\end{table}
\clearpage

\newpage
\section{Talk2Car-Expr Dataset Statistics}
\begin{figure}[htbp]
    \centering
    \includegraphics{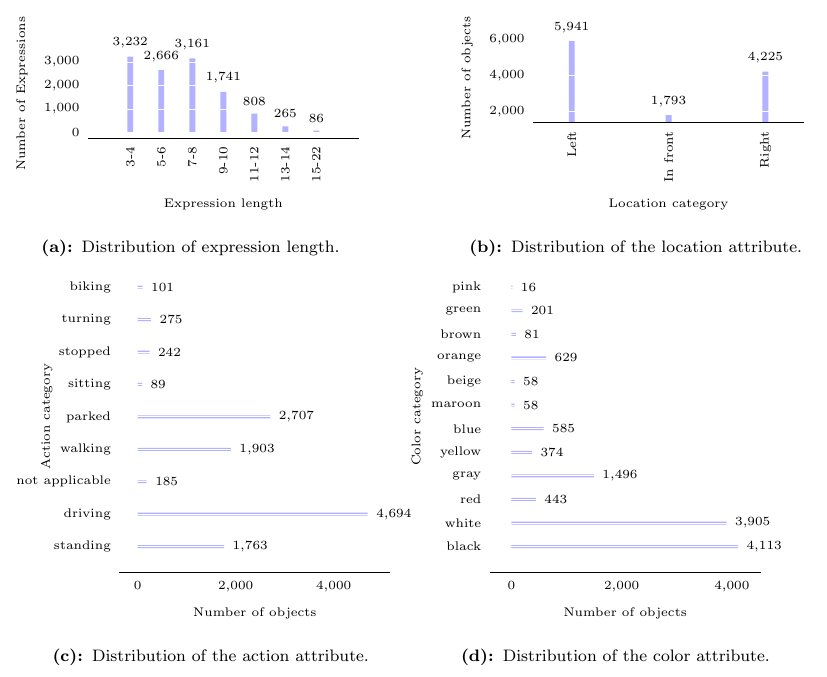}
    \caption{Dataset statistics of the Talk2Car-Expr dataset.}
    \label{fig:t2c_expr_data_statistics}
\end{figure}

In this section we show some of the statistics for the newly created Talk2Car-Expr dataset.
In {\color{black}Subfigure} \ref{fig:t2c_expr_data_statistics}(a) we show the distribution of the expression lengths, {\color{black}Subfigure} \ref{fig:t2c_expr_data_statistics}(b) shows the distribution of the location attributes, {\color{black}Subfigure} \ref{fig:t2c_expr_data_statistics}(c) shows the distribution of the action attributes, and {\color{black}Subfigure}  \ref{fig:t2c_expr_data_statistics}(d) shows the distribution of the color attributes.

\section{Generated Expressions Examples}
\label{app:examples_generated_expressions}
\begin{figure}[H]
\centering
\begin{subfigure}[t]{0.47\textwidth}
    \centering
    \includegraphics[width=\linewidth]{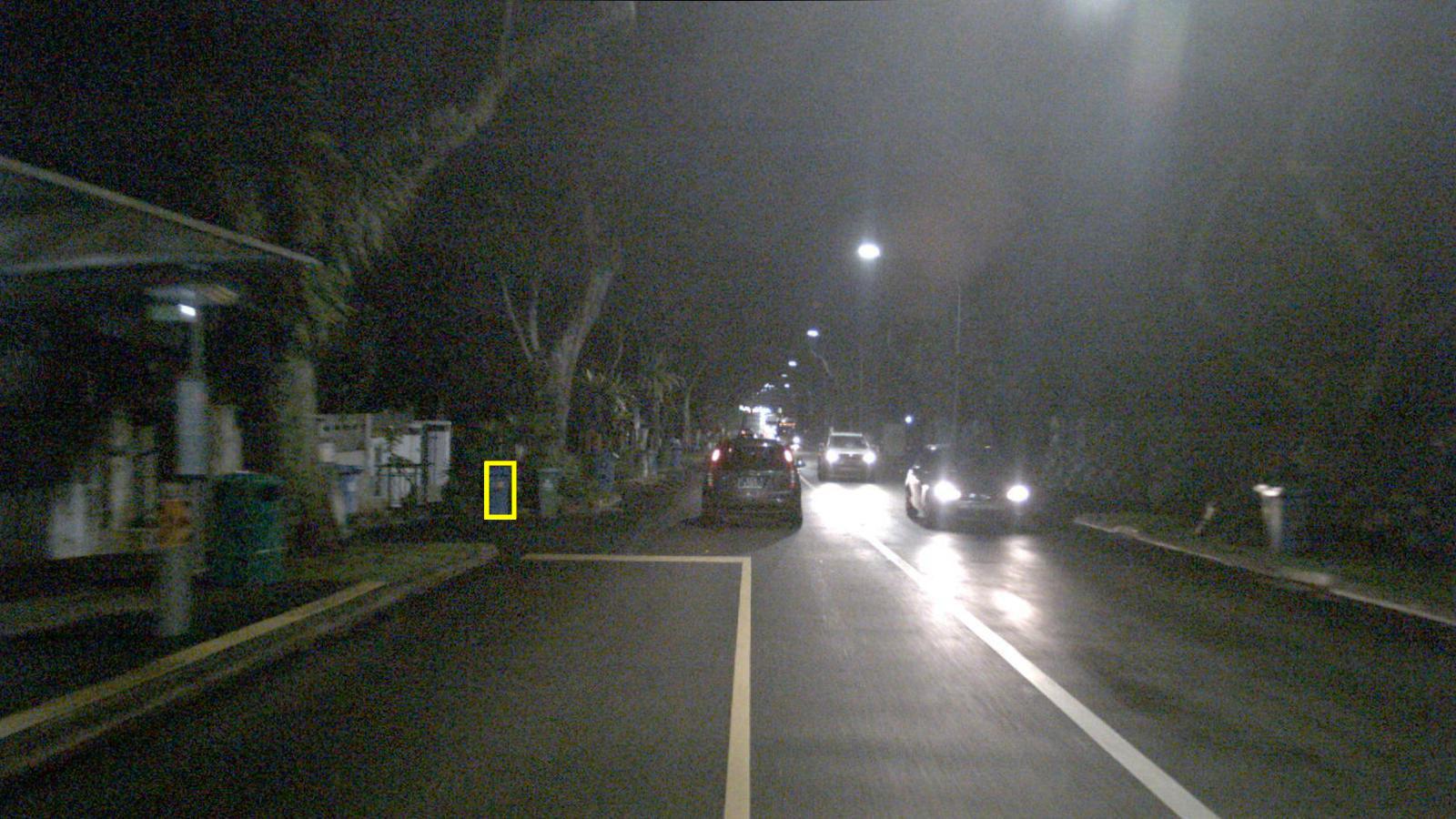}
\caption{Positive example for \gls{reg_model} models compared to baselines.}{\footnotesize
\begin{tabular}{rl}
    \textbf{SLR} & {\color{red}first pushable container in front}\\
    \textbf{SR} & {\color{darkgreen}first pushable container on left}\\
    \textbf{A-REG-att+cls+diff} & {\color{darkgreen}the first blue pushable container on the left}\\
    \textbf{A-REG-hot+cls+diff} & {\color{darkgreen}first pushable container on left}
\end{tabular}
}
    \label{fig:gen_expr_succes1}
\end{subfigure}
\begin{subfigure}[t]{0.47\textwidth}
    \centering
    \includegraphics[width=\linewidth]{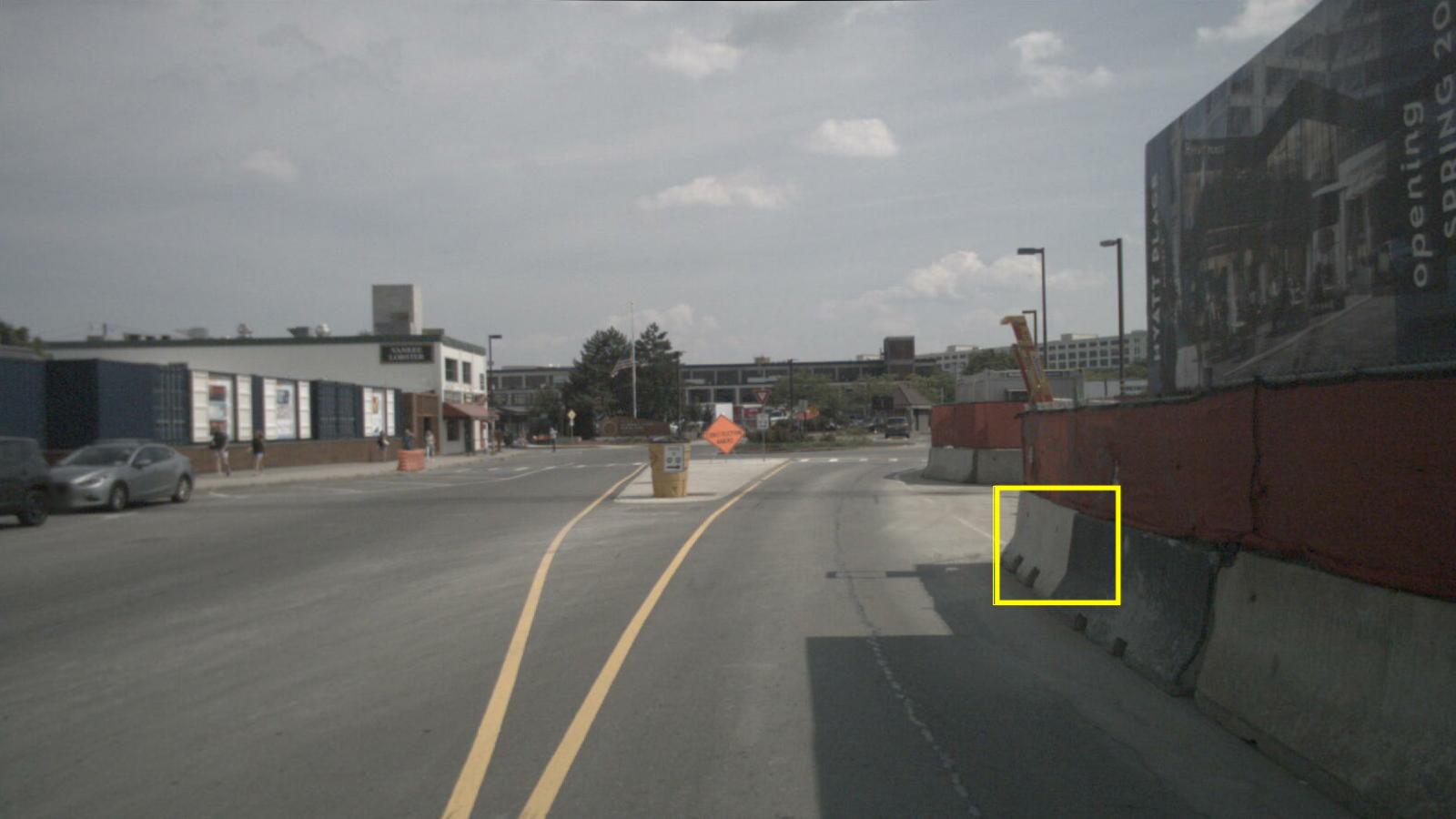}
\caption{\textcolor{black}{Positive example for \gls{reg_model}-att.
}}{\footnotesize
\begin{tabular}{rl}
    \textbf{SLR} & {\color{red}the first orange barrier on the right} \\
    \textbf{SR} & {\color{red}first barrier on right}\\
    \textbf{A-REG-att+cls+diff} & {\color{darkgreen}the first white barrier on the right}\\
    \textbf{A-REG-hot+cls+diff} & {\color{red}the first white barrier on the left}
\end{tabular}
}
    \label{fig:gen_expr_succes4}
\end{subfigure}
\begin{subfigure}[t]{0.47\textwidth}
    \centering
    \includegraphics[width=\linewidth]{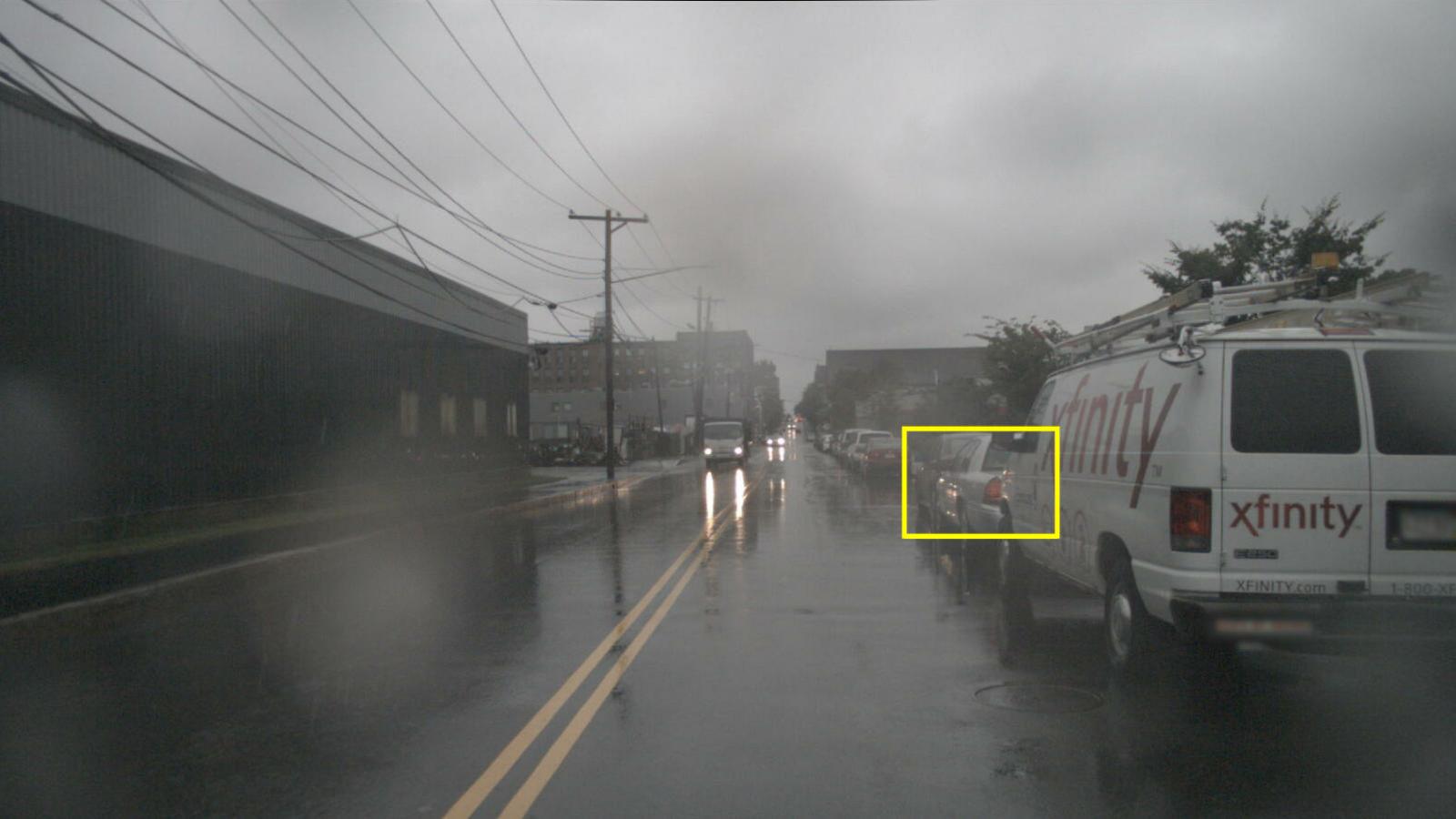}
\caption{Positive example for \gls{reg_model} compared to baselines.}{\footnotesize
\begin{tabular}{rl}
    \textbf{SLR} & {\color{red}the third white barrier parked on the right}\\
    \textbf{SR} & {\color{red}the third gray car parked on the right side of the street}\\
    \textbf{A-REG-att+cls+diff} & {\color{darkgreen}first white car on right}\\
    \textbf{A-REG-hot+cls+diff} & {\color{darkgreen}first car on right}
\end{tabular}
}
    \label{fig:gen_expr_succes6}
\end{subfigure}
\begin{subfigure}[t]{0.47\textwidth}
   \centering
    \includegraphics[width=\linewidth]{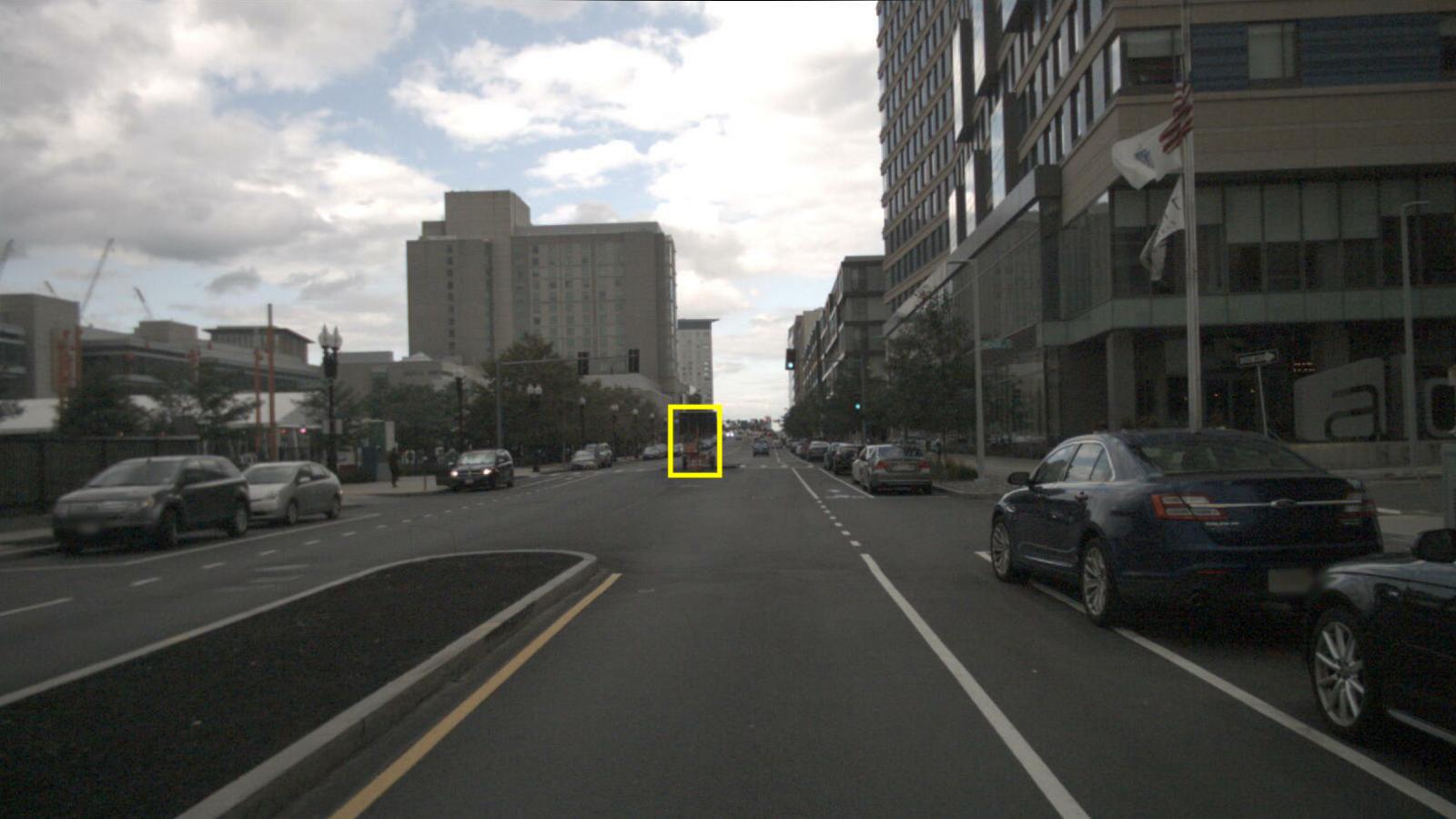}
     \caption{Negative example for all referring expression models.}{\footnotesize
    \begin{tabular}{rl}
        \textbf{SLR} & {\color{red}first construction worker in front}\\
        \textbf{SR} & {\color{red}first construction worker in front}\\
        \textbf{A-REG-att+cls+diff} & {\color{red}first trailer on left} \\
        \textbf{A-REG-hot+cls+diff} & {\color{red}first trailer on left}
    \end{tabular}
    }
    \label{fig:gen_expr_fail1}
\end{subfigure}
\begin{subfigure}[t]{0.47\textwidth}
\centering
    \centering
    \includegraphics[width=\linewidth]{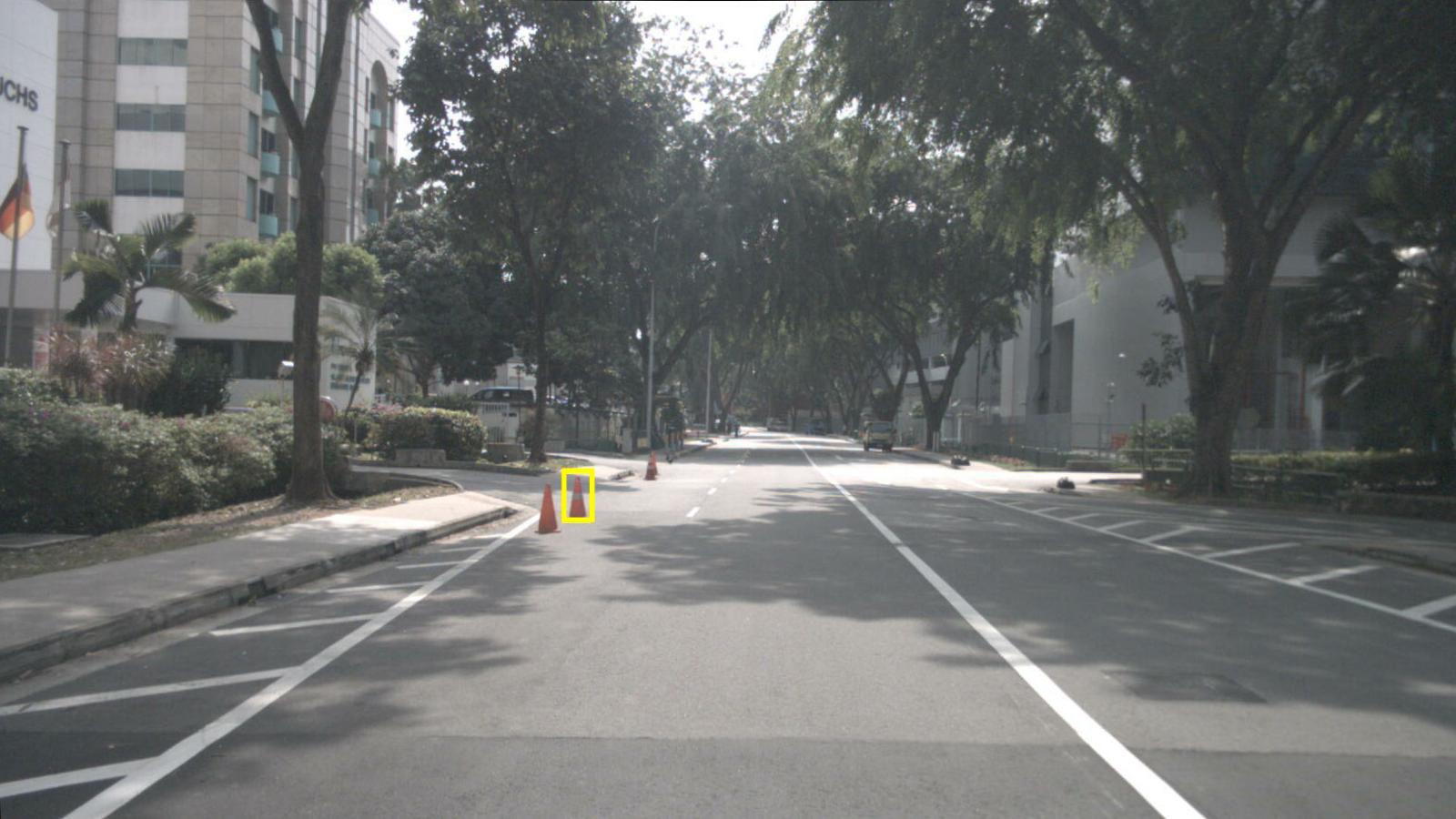}
\caption{Negative example for all referring expression models.}{\footnotesize
\begin{tabular}{rl}
    \textbf{SLR} & {\color{red}the first orange traffic cone in front of us}\\
    \textbf{SR} & {\color{red}first orange traffic cone in front} \\
    \textbf{A-REG-att+cls+diff} & {\color{red}the first orange traffic cone on the left} \\
    \textbf{A-REG-hot+cls+diff} &  {\color{red}the first orange traffic cone on the left}
\end{tabular}
}
    \label{fig:gen_expr_fail3}
\end{subfigure}
\begin{subfigure}[t]{0.47\textwidth}
\centering
    \centering
    \includegraphics[width=\linewidth]{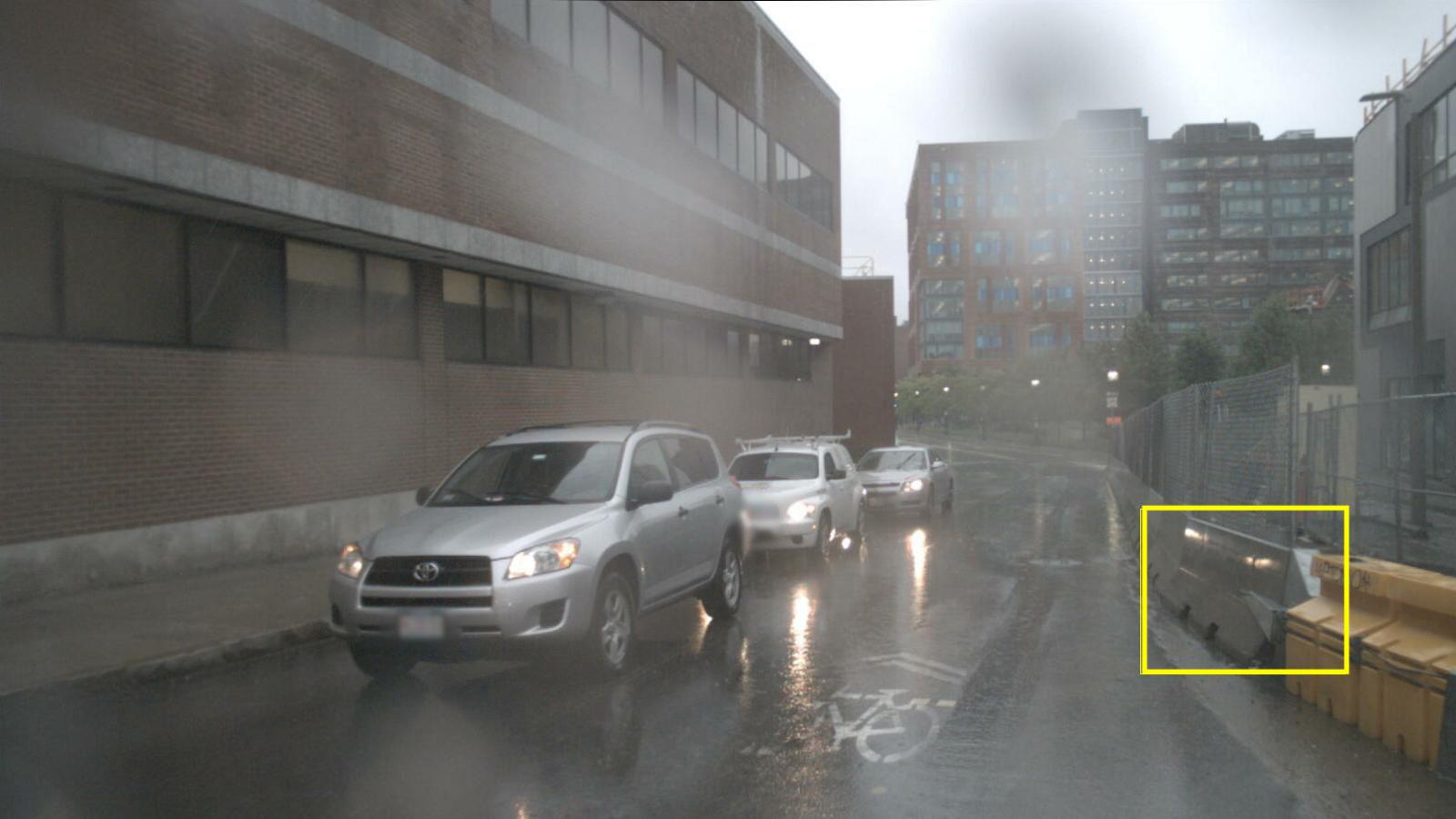}
\caption{Negative example except for SLR.}{\footnotesize
\begin{tabular}{rl}
    \textbf{SLR} & {\color{darkgreen}a white barrier on the right side}\\
    \textbf{SR} & {\color{red}\specialcell{the second gray truck parked on\\the right side of the street}}\\
    \textbf{A-REG-att+cls+diff} & {\color{red}the first orange barrier on the right} \\
    \textbf{A-REG-hot+cls+diff} & {\color{red}the first white barrier on the left}
\end{tabular}
}
    \label{fig:gen_expr_fail4}
\end{subfigure}
\caption{Examples from Expression Generation.}{Several examples for the expression generation for two of the baseline models (SLR and SR) and two variations of our model (\gls{reg_model}-att+cls+diff, \gls{reg_model}-hot+cls+diff). \textcolor{black}{We note that the \gls{reg_model}-hot tends to make mistakes in the position, i.e. predicting ``left'' when the object is on the right (subfigures (b) and (f)) and sometimes it leaves out the color from the generated expression (subfigures (a) and (c)). We assume the attention helps to focus on the importance of these attributes, which is missing in \gls{reg_model}-hot. We indicate in {\color{red}red} the wrong sentences and in {\color{darkgreen}green} the correct ones}
Best viewed in color}
\label{fig:all_appendix_examples}
\end{figure}
\end{document}